\documentclass{article} % For LaTeX2e
\usepackage{iclr2026_conference,times}

\usepackage{amsmath,amsfonts,bm}

\def\eqref#1{equation~\ref{#1}}
\def\1{\bm{1}}

\DeclareMathAlphabet{\mathsfit}{\encodingdefault}{\sfdefault}{m}{sl}
\SetMathAlphabet{\mathsfit}{bold}{\encodingdefault}{\sfdefault}{bx}{n}

\usepackage{hyperref}
\usepackage{url}
\usepackage{graphicx}
\usepackage{wrapfig}
\usepackage{algorithm}
\usepackage{algorithmic}
\usepackage{amsmath}
\usepackage{booktabs}
\usepackage{multirow}
\usepackage{siunitx}
\usepackage{graphicx}
\usepackage{makecell}
\usepackage[table, svgnames, dvipsnames]{xcolor}
\usepackage{comment}
\usepackage{tabularx}
\usepackage{booktabs}
\usepackage{subcaption}
\usepackage[table]{xcolor}

\newcolumntype{Y}{>{\centering\arraybackslash}X}

\title{WSVD: Weighted Low-Rank Approximation for Fast and Efficient Execution of Low-Precision Vision-Language Models}

\author{%
Haiyu Wang$^{1}$ \quad Yutong Wang$^{1}$ \quad Jack Jiang$^{2}$ \quad Sai Qian Zhang$^{1,2}$ \\
$^1$Tandon School of Engineering, New York University \\ $^2$Courant Institute of Mathematical Sciences, New York University\\
\texttt{\{hw3689,yw6594,jj2513,sai.zhang\}@nyu.edu}
}

\iclrfinalcopy % Uncomment for camera-ready version, but NOT for submission.
\begin{document}

\maketitle

\begin{abstract}
Singular Value Decomposition (SVD) has become an important technique for reducing the computational burden of Vision Language Models (VLMs), which play a central role in tasks such as image captioning and visual question answering. Although multiple prior works have proposed efficient SVD variants to enable low-rank operations, we find that in practice it remains difficult to achieve substantial latency reduction during model execution. To address this limitation, we introduce a new computational pattern and apply SVD at a finer granularity, enabling real and measurable improvements in execution latency. Furthermore, recognizing that weight elements differ in their relative importance, we adaptively allocate relative importance to each element during SVD process to better preserve accuracy, then extend this framework with quantization applied to both weights and activations, resulting in a highly efficient VLM. Collectively, we introduce~\textit{Weighted SVD} (WSVD), which outperforms other approaches by achieving over $1.8\times$ decoding speedup while preserving accuracy. We open source our code at: \href{https://github.com/SAI-Lab-NYU/WSVD}{\texttt{https://github.com/SAI-Lab-NYU/WSVD}}.
\end{abstract}

\section{Introduction}
\label{sec:intro}
Vision–language models (VLMs) represent a key frontier in artificial intelligence, as they connect visual recognition with natural language comprehension. By jointly processing imagery and text, these models enable diverse applications, including automatic image description~\citep{zhou2020unified, hu2022scaling, chen2022visualgpt, dzabraev2024vlrm}, visual question answering~\citep{chappuis2022prompt, bazi2023vision, wang2024surgical}, and semantic search over multimodal data~\citep{li2024searchlvlms,sun2025leveraging}. However, the impressive capabilities of VLMs come at the expense of significant resource demands. The joint encoding of large-scale visual and linguistic inputs requires heavy computation, and the autoregressive generation of tokens further stresses memory bandwidth, introducing major inference bottlenecks. 

To reduce the computational cost of large models, low-rank decomposition has recently attracted increasing attention~\citep{wang2025svdllmv2,yuan2023asvd,wang2024svd,li2025adasvd,li2024svdqunat,chang2024palu,wang2025dobi}. By factorizing the query (Q), key (K), and value (V) matrices within self-attention blocks into low-rank components, prior work has shown significant reductions in computational complexity and weight storage, thereby improving efficiency. However, based on our practical system-level implementation, we observe that applying SVD-based decomposition to the QKV matrices does not consistently yield latency improvements; in fact, it can sometimes incur even higher computational costs for some VLMs.

To investigate this, we first evaluate the latency of VLMs and find that the root cause lies in the recomputation of the KV vectors introduced by low-rank factorization, which requires multiple rounds of memory access to the latent data and ultimately increases memory traffic. To overcome this limitation, we propose a new computational pattern that applies SVD at a finer granularity, thereby achieving tangible and measurable improvements in execution latency.

Furthermore, building on prior work~\citep{yu2024super} demonstrating that certain weight elements play a critical role in VLM accuracy, we note that standard SVD operations treat all weights equally when truncating them for low-rank approximation. To address this, we adaptively allocate relative importance for each weight element during SVD to better preserve performance. 
To further enhance computational efficiency, we apply low-precision quantization to the low-rank VLM and finetune it to mitigate accuracy loss. Collectively, these steps yield a low-precision, low-rank VLM with exceptionally low execution latency. Our contributions are summarized as follows:
\begin{itemize}
    \item The WSVD scheme applies SVD separately to each attention head, fundamentally reducing memory access and computational cost during the decoding stage, and resulting in significantly lower VLM execution latency compared to prior solutions.
    \item To mitigate the accuracy drop introduced by the per-head SVD scheme, WSVD incorporates local weighted finetuning, where an importance score is assigned to each weight element during the SVD stage. This weighted decomposition produces low-rank weight matrices with minimal impact on VLM accuracy.
   \item WSVD applies quantization alongside SVD decomposition to both the weights and activations of the VLM. To further enhance efficiency, it incorporates an outlier elimination strategy within the SVD framework and locally finetunes the decomposed matrices, achieving improved accuracy while substantially reducing memory and computational cost.
\end{itemize}

\section{Related Work}
\label{sec:related-work}
\begin{figure}
    \centering
    \includegraphics[width=0.9\linewidth]{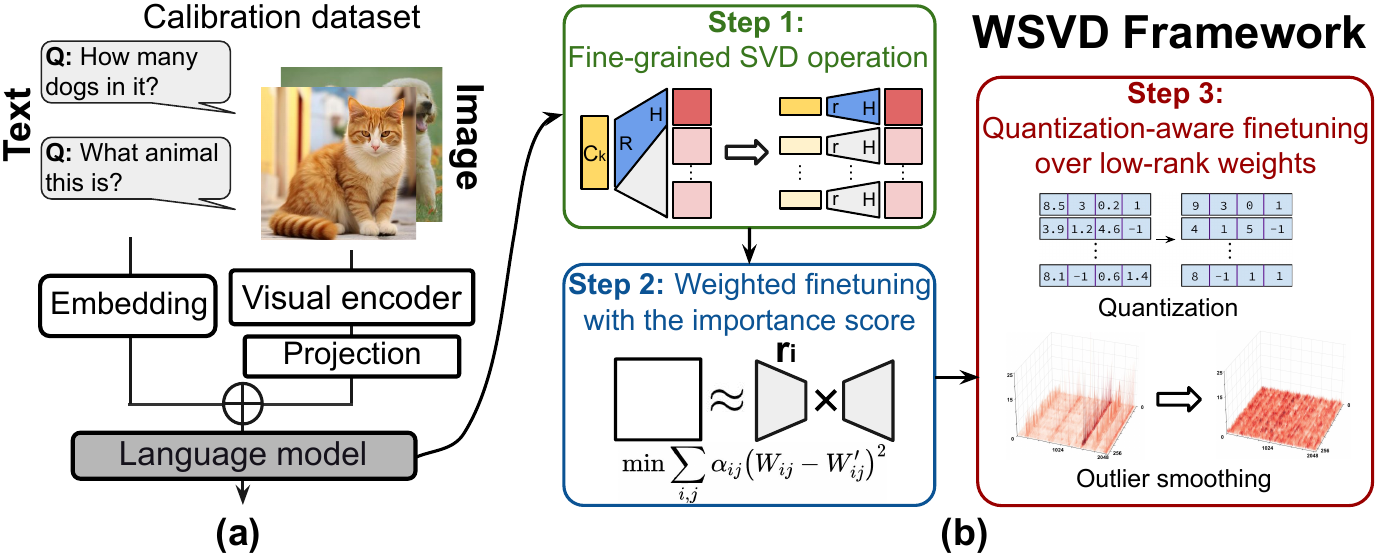}
    \vspace{-10pt}
    \caption{(a) Architecture of vision-language model. (b) Overview of WSVD framework.}
    \label{fig:wsvd_overall}
\end{figure}

\subsection{Vision Language Model}
\label{sec:related-vlm}
Vision–Language Models (VLMs)~\citep{li2022blip, li2023blip, liu2023visual-llava, 10.5555/3666122.3668264, beyer2024paligemma, grattafiori2024llama, wang2024qwen2} build on the foundation of Large Language Models (LLMs) by incorporating visual signals in addition to textual input, thereby enabling multimodal tasks such as image captioning and visual question answering (VQA). Representative systems like BLIP and InstructBLIP~\citep{li2022blip, li2023blip} leverage large-scale data curation and visual instruction tuning to better align their responses with human intent, particularly under zero-shot evaluation. A widely adopted framework, shown in Figure~\ref{fig:wsvd_overall} (a), encodes images into visual tokens via a vision backbone, concatenates them with text tokens, and feeds the combined sequence into a language model for output generation. This simple yet effective concatenation strategy underpins popular VLMs such as the LLaVA family~\citep{liu2023visual-llava}, SmolVLM~\citep{marafioti2025smolvlm}, PaLI-Gemma~\citep{beyer2024paligemma}, and Qwen-VL~\citep{wang2024qwen2}. Despite their strong performance, these models are often computationally heavy and difficult to deploy on devices with limited resources. To address this, compact designs have been introduced. TinyGPT-V~\citep{yuan2023tinygpt} and TinyLLaVA~\citep{zhou2024tinyllava} pursue scaled-down yet efficient alternatives, while SmolVLM~\citep{marafioti2025smolvlm} presents a family of lightweight models with one to three billion parameters that preserve competitive accuracy while significantly lowering hardware requirements.

\subsection{Singular Value Decomposition for Large Models}
\label{sec:related-svd}
Singular Value Decomposition (SVD)~\citep{jolliffe2016principal} is a fundamental tool in matrix factorization that represents a matrix $W \in \mathbb{R}^{m \times n}$ as $W = U \Sigma V^T$, where $U$ and $V$ are orthogonal matrices containing the left and right singular vectors, and $\Sigma$ is a diagonal matrix with non-negative singular values sorted in descending order. By retaining only the leading $r$ singular values and their associated vectors, one obtains a compact rank-$r$ approximation:
\begin{equation}
\label{eqn:svd}
W \approx U_r \Sigma_r V_r^T
\end{equation}
with $U_r \in \mathbb{R}^{m \times r}$, $\Sigma_r \in \mathbb{R}^{r \times r}$, and $V_r \in \mathbb{R}^{n \times r}$. This form can equivalently be written as $W \approx AB$, where $A = U_r \Sigma_r^{1/2}$ and $B = \Sigma_r^{1/2} V_r^T$. Such low-rank approximations capture the dominant structure of $W$, allowing dimensionality reduction, compression, and faster computation.
SVD has been extensively studied as a compression strategy for LLMs~\citep{wang2025svdllmv2,yuan2023asvd,wang2024svd,li2025adasvd,li2024svdqunat,chang2024palu,wang2025dobi}. 
Early work~\citep{noach2020compressing} applied vanilla SVD directly to weight matrices, but the method suffered from considerable approximation errors. Subsequent techniques refined this approach: FWSVD~\citep{hsu2022language} incorporates Fisher information~\citep{ly2017tutorial} to rank parameter importance, ASVD~\citep{yuan2023asvd} accounts for activation outliers, and SVD-LLM~\citep{wang2024svd} explicitly minimizes the loss introduced by discarded singular values. 

While most efforts have focused on compressing model weights, it can also be used for KV cache compression~\citep{chang2024palu, yu2024effectively}.
In particular, the key and value projection matrices can be factorized as $W_{K} = A_{K}B_{K}$ and $W_{V} = A_{V}B_{V}$. For a given input $X$, this allows the KV cache to store only the low-dimensional latent vectors $C_{K} = XA_{K}$ and $C_{V} = XA_{V}$, thereby reducing cache size. During decoding, the original KV representations are reconstructed via $K = C_{K}B_{K}$ and $V = C_{V}B_{V}$.
More recent innovations include AdaSVD~\citep{li2025adasvd}, which dynamically adjusts compression rates based on per-layer sensitivity, and SVD-LLM2~\citep{wang2025svdllmv2}, which optimizes truncation strategies using theoretical error analysis.

%Complementary to these static factorization methods, DeepSeek introduces Multi-Head Latent Attention (MLA)~\citep{liu2024deepseek}, which integrates low-rank projections directly within the attention mechanism. Instead of operating on full key and value matrices, MLA projects them into compact latent representations through learned mappings, thereby reducing both memory consumption and computation during inference. This approach can be interpreted as an implicit low-rank factorization applied on the fly, offering performance gains that synergize with traditional SVD-based compression techniques.

\subsection{Fisher-Based Importance and Weighted Matrix Factorization}
\label{sec:related-importance}
Fisher information has been widely used as a measure of parameter importance in continual learning~\citep{kirkpatrick2017overcoming} and in pruning and compression~\citep{liu2021groupfisher,singh2020woodfisher}. Weighted low-rank approximation has been explored in matrix completion and recommendation, where each entry carries a confidence weight~\citep{srebro2003weighted}. More recently, FWSVD~\citep{hsu2022language} incorporates Fisher information into low-rank factorization by assigning a single Fisher-based weight to each row and applying SVD to a pre-scaled matrix, yielding a coarse row-wise weighting. On the interpretability side, gradient-based attribution and layer-wise relevance propagation methods~\citep{ancona2017towards,bach2015pixel} also use importance weights, but primarily for explanation rather than compression. In contrast, WSVD uses element-wise Fisher weights to directly guide both local fine-tuning and quantization-aware training.

\subsection{Flash Decoding}
\label{sec:related-fd}
FlashAttention~\citep{dao2022flashattention} is an IO-aware attention algorithm that leverages tiling and kernel fusion to reduce memory traffic and improve GPU utilization. 
By keeping query, key, and value tiles in on-chip memory and streaming them through a fused kernel, FlashAttention avoids materializing large intermediate attention matrices, thereby lowering memory footprint and achieving substantial speedups in training and inference. 

Building on this idea, Flash Decoding~\citep{dao2023flashdecoding} extends FlashAttention to the autoregressive decoding setting. 
Instead of materializing and reloading the entire KV cache for each step, it streams $K$ and $V$ in sequence tiles and incrementally updates online softmax statistics. 
This block-wise processing exposes additional parallelism along the sequence length dimension, complementing the existing head- and batch-level parallelism in FlashAttention, and thereby improves GPU utilization. 
As a result, Flash Decoding achieves both lower memory traffic and higher throughput, and has become the de facto baseline for efficient inference in large language and vision-language models. 
Our WSVD system further builds on Flash Decoding by integrating low-rank reconstruction into the fused kernel pipeline, reducing memory overhead while preserving its efficiency (see Section~\ref{sec:wsvd-sys}).

\section{Method}
An overview of WSVD is presented in Figure~\ref{fig:wsvd_overall} (b), which consists of three key components: (i) Per-head SVD operations for reduced latency (Section~\ref{sec:per-head-svd}), (ii) WSVD with dynamic importance scoring (Section~\ref{sec:wsvd}), and (iii) quantization-aware finetuning for low-rank VLMs (Section~\ref{sec:qat}).

\subsection{Fine-grained Per-Head SVD Operation for Reduced Latency}
\label{sec:per-head-svd}
%\textcolor{red}{Haiyu: present a experimental study on GPU, show the latency breakdown for a single layer, and memory access for the conventional solution, illustrate the Reason, discuss Fine-Grained Low-Rank Decomposition for Memory-Efficient VLM Execution}

Prior studies have shown that VLM decoding is predominantly memory-bound, as long image-token sequences enlarge the KV cache and each generated token requires accessing the large KV cache, with overall latency bottlenecked by memory access. As discussed in Section~\ref{sec:related-svd}, conventional SVD-based approaches~\citep{chang2024palu, wang2025qsvd, wang2024svd} address this by decomposing projection matrices (e.g., $Q$, $K$, and $V$), thereby reducing parameter count and storing low-rank latent representations $C_{K}$ and $C_{V}$. This strategy not only decreases computation and runtime in the prefill stage but also reduces cache size, offering potential I/O savings during the decoding stage.

\begin{figure}
    \centering
    \includegraphics[width=\linewidth]{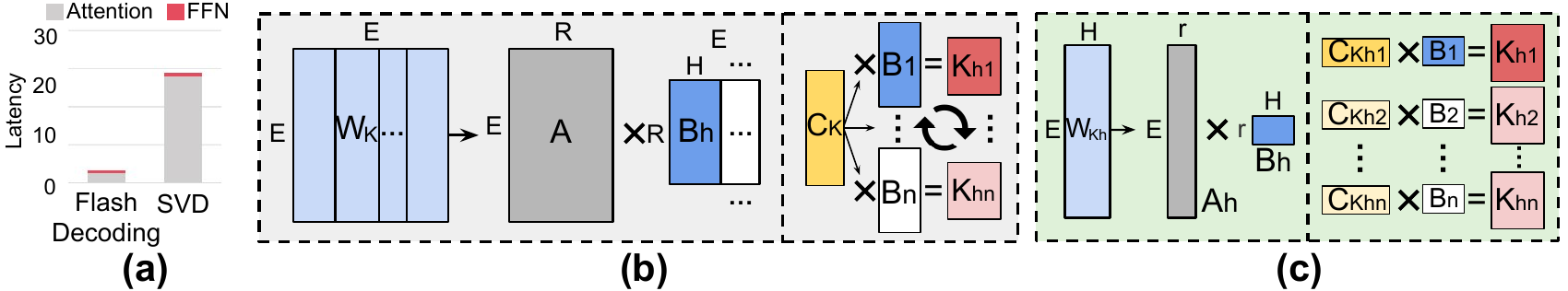}
    \vspace{-15pt}
    \caption{
    (a) Latency evaluation of VLM including self-attention (SA) and feed-forward (FFN) modules. 
    (b) Conventional SVD: the left side illustrates SVD of $W_k$, and the right side shows the reconstruction of $K_h$ from the shared latent. 
    (c) Per-head SVD: the left side illustrates per-head SVD of $W_{Kh}$, and right side shows per-head reconstruction of $K_h$ from per-head latent.
    }
    \label{fig:per-head-svd}
    \vspace{-10pt}
\end{figure}

However, in practice, we find that reconstructing $K$ and $V$ from low-rank latents introduces substantial overhead, leading to even higher decoding latency than the original uncompressed model. Specifically, 
we profile the single-layer decoding latency of LLaVA-Next 7B~\citep{zhou2024tinyllava} on an RTX~4090, comparing standard flash decoding without SVD against an SVD baseline that caches low-rank latents. In this baseline, both the rank ratio and cache size are reduced to 50\% as before. %and decoding is implemented using our fused reconstruction and flash decoding kernel as described in Section~\ref{sec:wsvd-sys}. 
With a batch size of 16 and a KV cache length of 8192, the results (Figure~\ref{fig:per-head-svd} (a)) show that SVD scheme incurs substantially higher latency compared to flash decoding.

To pinpoint the cause of this latency growth, we observe that the overhead arises from decomposing the entire $K$ and $V$ matrices. Taking $K$ as an example, after SVD we obtain $W_{K} = A_{K}B_{K}$, where $A_{K} \in \mathbb{R}^{E \times R}$ and $B_{K} \in \mathbb{R}^{R \times E}$, with $E$ denoting the embedding dimension and $R$ the truncated rank. For each head $h$, the key projection is computed as $W_{Kh} = A_{K}B_{Kh}$, where $B_{Kh} \in \mathbb{R}^{R \times H}$ and $H$ is the head dimension (Figure~\ref{fig:per-head-svd}(b)). During inference, the latent representation $C_{K} = XA_{K} \in \mathbb{R}^{L \times R}$ is cached across sequence length $L$, and each head’s key vector is reconstructed as $K_{h} = C_{K}B_{Kh}$. This reconstruction introduces a computational cost of $\gamma_{\text{svd}} = LRH$ per head. Compared with directly storing the $K$ matrix of size $LE$, although caching $C_{K}$ reduces storage to $LR$,~\textbf{reconstructing $W_{Kh}$ for each head requires accessing the entire $C_{K}$}, which has a size $LR$. As a result, the effective memory footprint becomes $\eta_{\text{svd}} = LR$ per head, thereby increasing decoding latency. Similar argument holds trues for the computation of value vector $V$.

To mitigate this overhead, our WSVD approach applies SVD directly to the submatrices of $W_{K}$ and $W_{V}$ corresponding to each head, rather than decomposing the entire matrices, as illustrated in Figure~\ref{fig:per-head-svd} (c). Specifically, for head $h$, the submatrix $W_{Kh} \in \mathbb{R}^{E \times H}$ is decomposed as $W_{Kh} = A_{Kh} B_{Kh}$, where $A_{Kh} \in \mathbb{R}^{E \times r}$ and $B_{Kh} \in \mathbb{R}^{r \times H}$. The rank $r$ is obtained by truncating the $H$ singular values of $W_{Kh}$. Since $H \ll E$, the per-head rank $r$ is typically much smaller than $R$. 
For each head $h$, the latent representation is computed as $C_{Kh} = X A_{Kh} \in \mathbb{R}^{L \times r}$ and stored in the cache. During decoding, the corresponding key vector is reconstructed as $K_h = C_{Kh} B_{Kh}$. Unlike the conventional SVD approach shown in Figure~\ref{fig:per-head-svd} (b), this design eliminates the need to repeatedly load a large shared latent representation $C_K$, since ~\textbf{each head can be reconstructed directly from its own latent $C_{Kh}$}. With this design, the memory footprint is reduced to $\eta_{\text{wsvd}} = Lr$, since only the latent vector $C_{Kh}$ needs to be stored, and the computational cost of reconstructing $K_h$ becomes $\gamma_{\text{wsvd}} = LrH$, where $r \ll R$. A similar computation applies to the reconstruction of $V$.

To evaluate the saving analytically, the per-head SVD scheme shown in Figure~\ref{fig:per-head-svd} (c) reduces both memory traffic and computational cost, thereby enabling practical decoding acceleration, as demonstrated in Section~\ref{sec:eval-system}. In particular,
\begin{equation}
\label{eq:per-head-full}
\small
\frac{\gamma_{\text{wsvd}}}{\gamma_{\text{svd}}} = \frac{\eta_{\text{wsvd}}}{\eta_{\text{svd}}} = \frac{r}{R}, 
\quad r \ll R.
\end{equation}
Thus, both computational cost and memory footprint for latent storage are reduced by a factor of $r/R$. Compared to the original SVD-based scheme, WSVD further reduces the weight parameter count from $\alpha_{\text{orig}} = EH$ per head to $\alpha_{\text{wsvd}} = Er + rH$, and lowers the KV-cache size from $\eta_{\text{orig}} = LH$ to $\eta_{\text{wsvd}} = Lr$. These improvements are quantified by the parameter size ratio $\rho_1$ and the cache size ratio $\rho_2$ for KV vector storage.
\begin{equation}
\label{eq:per-head-orig}
\small
\rho_1 = \frac{\alpha_{\text{wsvd}}}{\alpha_{\text{orig}}} = \frac{(E+H)\times r}{E \times H}=(1+\frac{H}{E})\cdot\frac{r}{H} , 
\qquad 
\rho_2 =\frac{\eta_{\text{wsvd}}}{\eta_{\text{orig}}} = \frac{r}{H}.
\end{equation}
However, the per-head SVD in the WSVD scheme also amplifies approximation errors, making accuracy degradation harder to control compared to conventional SVD applied to the full $W_k$. Next, we describe a local weighted finetuning scheme to mitigate the accuracy drop.

\subsection{SVD with Local Weighted Finetuning}
\label{sec:wsvd}

% Figure~\ref{fig:gateup_grad_ana} illustrates this phenomenon for the LLaVA-1.5 model, where gradient analysis of the FFN Gate/Up projections reveals the distinctive importance of each projection. 

% \begin{wrapfigure}{r}{0.5\textwidth}
%     \centering
%     \includegraphics[width=0.5\columnwidth]{fig/gateupimg.png}
%     \caption{Gradient analysis on FFN gate/up projection.}
%     \label{fig:gateup_grad_ana}
% \end{wrapfigure}
Conventional SVD converts a full-rank input matrix into a low-rank representation, but one limitation is that it cannot control the relative contribution of different weights after decomposition. Prior work~\citep{yu2024super} has shown that in large models, weights vary significantly in their importance to final accuracy. In particular, some ``superweights'' are highly sensitive, where even small changes in magnitude can cause a substantial drop in accuracy. Therefore, it is crucial to incorporate this notion of importance when performing SVD, resulting in a weighted low-rank decomposition.

The first question is how to evaluate the importance of a weight element. To formalize this, let $\mathcal{D}$ denote the data distribution over calibration samples $x$, and let $\ell(W;x)$ denote the training loss of sample $x$. The importance score of each element in $W_{K}$ with respect to final accuracy can be estimated as:
\begin{equation}
    G_K = \mathbb{E}_{x \sim \mathcal{D}}\big[\nabla_{W_K}\,\ell(W; x)\big].
\end{equation}
A weight entry with a large gradient magnitude indicates that even a small change in this element has a substantial effect on the expected model loss. Accordingly, $G_K$ can be interpreted as an importance score that links parameter updates to their impact on performance.

This estimation of training loss impact can be refined using the Fisher Information Matrix (FIM), which quantifies parameter importance as the expected sensitivity of the log-likelihood with respect to model parameters. A second-order Taylor expansion of the expected loss around the current parameter values yields:
\begin{align}
\small
\Delta \mathcal{L} 
&= \mathbb{E}_{x\sim\mathcal{D}}\!\big[\ell(W+\Delta W; x) - \ell(W; x)\big] \\
% &\approx \big\langle \, \mathbb{E}_{x}\!\big[\nabla_W \ell(W; x)\big],\, \Delta W \big\rangle
% + \tfrac{1}{2}\, \Delta W^\top 
% \Big(\mathbb{E}_{x}\!\big[\nabla_W^2 \ell(W; x)\big]\Big) \Delta W .
&\approx 
% \Delta W^\top \, \mathbb{E}_{x}\!\big[\nabla_W \ell(W; x)\big]
% + 
\tfrac{1}{2}\, \Delta W^\top 
\Big(\mathbb{E}_{x}\!\big[\nabla_W^2 \ell(W; x)\big]\Big) \Delta W .
\end{align}
%\end{equation}
To make the computation of the Hessian tractable, it can be approximated by a diagonal matrix, where each diagonal entry corresponds to the Fisher importance score of the parameter. For example, the vector of Fisher information score $F_K$ for $W_{K}$ can be computed as:
\begin{equation}
F_K = \mathbb{E}_{x\sim\mathcal{D}}\!\big[g_K(x) \odot g_K(x)\big], \;\;\;g_K(x) =\nabla_{W_K}\,\ell(W; x)
\end{equation}
where $\odot$ denotes elementwise multiplication. Motivated by these observations, we propose a weighted local fine-tuning mechanism that performs SVD while incorporating the relative importance of each weight element, quantified by its Fisher information score. Specifically, the objective function can be described as:
\begin{equation}
    \small
    \min_{A_{K}, B_{K}}\bigl\| F_{K}^{1/2} \odot (W_{K} - A_{K} B_{K}) \bigr\|_F^2
\end{equation}
where $A_{K}$, $B_{K}$ are the low-rank decomposition to estimate $W_{K}$. In the context of per-head SVD described in Section~\ref{sec:per-head-svd}, this optimization can be applied across the SVD for the weight matrices for each head $h$, and the objective function can be depicted as:
\begin{equation}
    \label{eqn:wsvd_loss}
    \small
    \min_{A_{Kh}, B_{Kh}}\sum_{h} \bigl\| F_{Kh}^{1/2} \odot (W_{Kh} - A_{Kh} B_{Kh}) \bigr\|_F^2
\end{equation}
where $A_{Kh}$ and $B_{Kh}$ denote the low-rank approximation of $W_{Kh}$. Since no analytical solution exists for this problem, it is solved by fine-tuning $A_{Kh}$ and $B_{Kh}$ until convergence. The same loss formulation can be applied to other projection matrices in the model (e.g., $W_Q$, $W_V$, or feed-forward layers), providing a general framework for gradient-weighted fine-tuning after SVD truncation. The WSVD procedure is summarized in Algorithm~\ref{alg:wsvd}.

\subsection{Local Quantization-aware Training for Low-Precision WSVD}
\label{sec:qat}
To further reduce model size and cache footprint, we apply low-precision quantization to the low-rank model parameters and the input and mitigate accuracy loss using local quantization-aware training (QAT). To address channel-wise outliers in the input $X$ and latent representations $C_{K}, C_{V}$, we follow previous work~\citep{ashkboos2024quarot,xiang2024dfrot} and introduce two orthogonal matrices $S_{1}$ and $S_{2}$, and $S_{1}$ is also a Hadamard matrix with predefined binary elements. With these transformations, the quantized $Q, K, V$ computation for each head $h$ can be reformulated as: % and its quantized counterpart
\begin{equation}
\label{eqn:quant-had}
Y_h = X A_h B_h \rightarrow Y_h = (X S_{1}^{\top})(S_{1} A_h S_{2}^{\top})(S_{2} B_h) \approx Q(X S_{1}^{\top})Q(S_{1} A_h S_{2}^{\top})Q(S_{2} B_h)
\end{equation}
where $S_{1}^{\top}S_{1} = S_{2}^{\top}S_{2} = I$, $Q(\cdot)$ denotes the quantization operator, and we omit the QKV subscripts for simplicity of presentation. We further finetune the rotational matrices $S_{2}$ together with $A_{h}, B_{h}$ to minimize the change on the low-rank weights caused by quantization, with the objective as follows:
\begin{equation}
\label{eqn:qat-loss}
\small
\min_{S_{2}, A_{h}, B_{h}}\bigl\|\, (F'_h)^{1/2} \odot \bigl[ S_1 W_h - Q(S_1 A_h S_2^{\top}) \, Q(S_2 B_h) \bigr] \,\bigr\|_2
\end{equation}
where $F'_h \;\approx\; \mathbb{E}_{x\sim\mathcal{D}} [(S_1g(x)) \odot (S_1g(x))]$. $F'_h$ is the Fisher information matrix associated with the transformed weight $S_1 W_h$, computed element-wise as the root of the expected squared gradient $S_1g(x)$ over the calibration dataset $\mathcal{D}$. This acts as an importance weight, emphasizing parameters with higher sensitivity and guiding the QAT objective to more effectively preserve accuracy under quantization. During QAT, we jointly update $A_h$, $S_2$, and $B_h$, while $S_1$ is fixed as an exact Hadamard matrix of size $E \times E$, determined by the model embedding dimension $E$. This update design enables the factorized components to flexibly adapt to quantization noise while preserving the orthogonal transformation imposed by $S_1$, thereby maintaining the low-rank structure and improving the approximation accuracy and stability of low-precision training. Since this procedure is QAT performed locally, it incurs much lower time and memory overhead than end-to-end finetuning.

\subsection{WSVD System Implementation}
\label{sec:wsvd-sys}
\begin{figure}
    \centering
    \includegraphics[width=\linewidth]{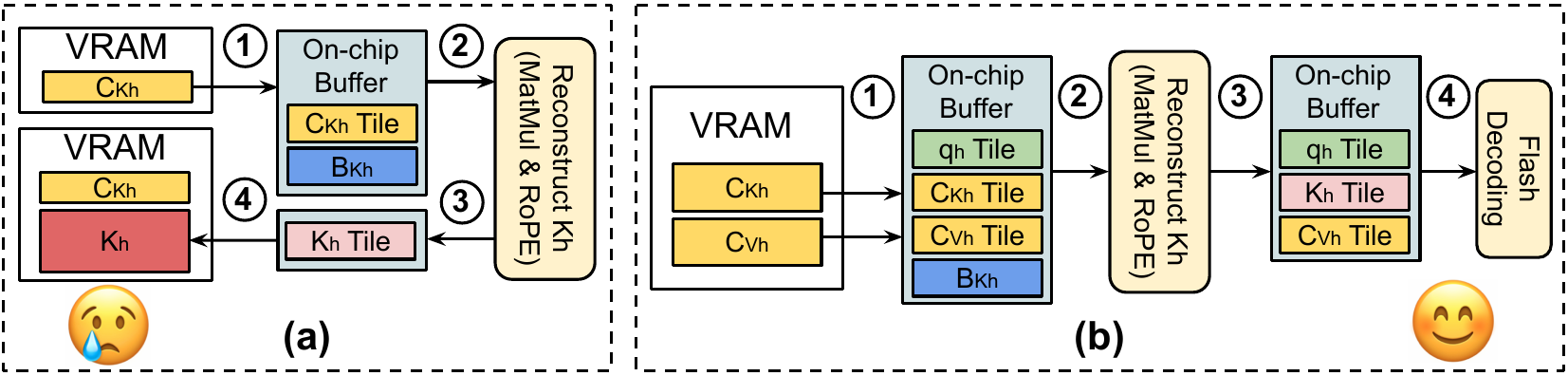}
    \vspace{-15pt}
    \caption{(a) Naive reconstruction requires materializing and writing back full $K_h$ to VRAM (global GPU memory), leading to excessive memory usage and I/O.  (b) Our fused kernel consumes $C_{Kh}$ and $C_{Vh}$ tiles on-chip with flash decoding, reducing both peak memory footprint and I/O traffic. All the step numbers are shown in circle.}
    \label{fig:system-impl}
    \vspace{-10pt}
\end{figure}

A naive PyTorch implementation of WSVD results in excessive memory consumption during the reconstruction of $K$ and $V$, as illustrated in Figure~\ref{fig:system-impl} (a). Taking the key $K_h$ of head $h$ as an example, with $K_h = C_{Kh}B_{Kh}$ where $C_{Kh}\in\mathbb{R}^{L\times r}$ and $B_{Kh}\in\mathbb{R}^{r\times H}$, the GPU operation proceeds as follows. First, the latent representation $C_{Kh}$ is loaded from VRAM. Next, reconstruction $C_{Kh}B_{Kh}$ is performed, materializing the full $K_h\in\mathbb{R}^{L\times H}$ in VRAM. The reconstructed $K_h$ is then written back to VRAM and later reloaded for attention. Since $K_h$ and $V_h$ cannot fit into limited on-chip buffers, they must be stored along with the latent $C_{Kh}, C_{Vh}$, which largely increases I/O traffic and peak memory usage, in some cases exceeding that of the original model without low-rank decomposition.

%A naive PyTorch implementation of WSVD leads to excessive memory consumption for the reconstruction of $K$ and $V$, as illustrated in Figure~\ref{fig:system-impl} (a). Taking the key $K_h$ of head $h$ as an example, where $K_h = C_{Kh} B_{Kh}$ with $C_{Kh}\in\mathbb{R}^{L\times r}$ and $B_{Kh}\in\mathbb{R}^{r\times H}$. For long sequences, the GPU operation proceeds as follows: (1) the latent representation $C_{Kh}$ is loaded from VRAM (GPU memory); (2) reconstruction $C_{Kh}B_{Kh}$ is performed, materializing the full $K_h\in\mathbb{R}^{L\times H}$ in VRAM; and (3–4) the reconstructed $K_h$ is written back to VRAM and later reloaded for attention. As $K_h$ and $V_h$ cannot fit into limited on-chip buffers, they must be stored alongside the latent $C_{Kh}, C_{Vh}$, which substantially increases I/O traffic and peak memory usage, even exceeding that of the original model without applying low-rank decomposition.

To address this problem, we design a fused kernel in Triton~\citep{tillet2019triton} that integrates low-rank reconstruction directly into the flash decoding pipeline, as shown in Figure~\ref{fig:system-impl} (b). 
At tile granularity, the kernel streams a tile $t$ of $C_{Kh}$, denoted $C_{Kh,t} \in \mathbb{R}^{l \times r}$, from VRAM (step 1), where $l$ is the tile size along the sequence dimension $L$ that fits into on-chip memory. The up-projection weight $B_{Kh}$ is then loaded once into on-chip storage (step 2), and the temporary key tile $K_{h,t} = C_{Kh,t} B_{Kh}$ is formed in registers or shared memory (step 3). This process is executed within a single fused kernel that proceeds directly into the flash decoding pipeline: the temporary $K_{h,t}$ is immediately contracted with the query tile $q_h$ to compute $q_h K_{h,t}^{\top}$, update the online softmax statistics, and apply the normalized attention weights to the corresponding value tile $C_{Vh,t}$ (step 4). 

In this design, both $C_{Kh}$ and $C_{Vh}$ are consumed in place, and all intermediate tensors remain on-chip without being written back to VRAM. The fused kernel integrates reconstruction, $qK^{\top}$ accumulation, softmax normalization, and the $V$ multiplication into a single workflow, eliminating redundant kernel launches and memory transfers. Memory usage now scales only with the tile size ($l \times r$ and $B_{Kh}$), which significantly reduces peak footprint and I/O traffic while preserving the efficiency of flash decoding. The design exposes parallelism at two levels: across tiles, where multiple tiles are processed concurrently within each head, and across heads, where different heads execute in parallel, fully utilizing GPU compute resources in accordance with flash decoding scheduling. Finally, the $V$-path up-projection $B_{Vh}$ is fused into the output projection, which avoids explicit reconstruction of $V_h$, following Palu~\citep{chang2024palu}. Collectively, these optimizations eliminate redundant memory operations while maintaining high parallelism, enabling WSVD to achieve practical inference acceleration without any loss of accuracy.

\begin{wrapfigure}{r}{0.45\linewidth}
    \vspace{-12pt}
    \centering
    \includegraphics[width=\linewidth]{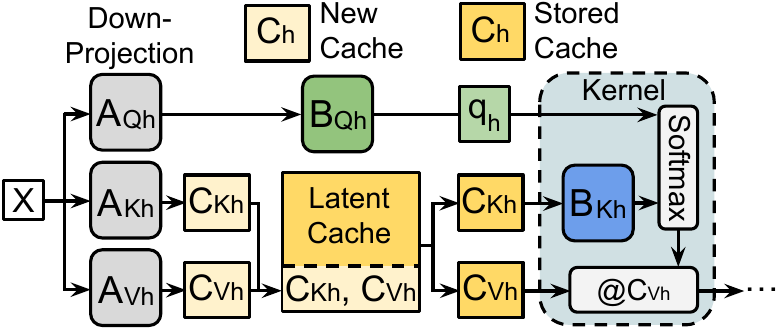}
    \vspace{-15pt}
    \caption{WSVD decoding pipeline. Each token is down-projected to low-rank latents, and $K$ and $V$ latents are appended to the cache, while $Q$ latent is up-projected and consumed together with cached $C_{Kh}, C_{Vh}$ in the fused kernel.}
    \label{fig:decoding-pipeline}
    \vspace{-10pt}
\end{wrapfigure}

Beyond kernel fusion, WSVD applies per-head SVD to the Query, Key, and Value projections to reduce parameters and improve efficiency. Decomposing $W_K$ and $W_V$ decreases model size and accelerates both prefilling and decoding, while decomposing $W_Q$ further reduces parameters and speeds up prefilling. During prefilling, the input sequence is projected into low-rank $Q,K,V$ latents, with $K,V$ latents stored as cache.

During the decoding stage, as shown in Figure~\ref{fig:decoding-pipeline}, each new token is processed through per-head down-projections to generate low-rank latents for $Q$, $K$, and $V$. The latents of $K$ and $V$ are stored in the cache, while the latent of $Q$ is immediately up-projected to form $q$ for the current attention step. The kernel then loads the cached latents $C_{Kh}$ and $C_{Vh}$ together with the current $q_h$, performing highly parallelized computation that integrates low-rank reconstruction with flash decoding. This unified pipeline eliminates redundant materialization of full $K$ and $V$, preserves compact latent caches throughout decoding, and enables efficient attention computation with a reduced memory footprint.

%During decoding stage, as illustrated in Figure~\ref{fig:decoding-pipeline}, each new token is similarly processed through per-head down-projections to produce low-rank $Q$, $K$, and $V$ latents. 
%The $K$ and $V$ latents are appended to the cache, whereas the $Q$ latent is immediately up-projected to form $q$ for the current attention step. 
%The kernel then loads the cached latent $C_{Kh}, C_{Vh}$ together with the current $q_h$ and performs highly parallelized computation that integrates low-rank reconstruction with flash decoding. 
%This unified pipeline avoids redundant materialization of full $K$ and $V$, maintains compact latent caches throughout decoding, and achieves efficient attention computation with reduced memory footprint.

\section{Evaluation}
\label{sec:eval}
We conduct experiments on five representative vision–language models: LLaVA-v1.5 7B~\citep{liu2023visual-llava}, LLaVA-v1.5 13B, LLaVA-Next 7B, LLaVA-Next 13B, and SmolVLM-Instruct~\citep{marafioti2025smolvlm}. For local weighted fine-tuning and QAT, we use 256 samples randomly drawn from the ScienceQA training split~\citep{lu2022learn_sqa}, following the procedures described in Section~\ref{sec:wsvd} and Section~\ref{sec:qat}. Evaluation is conducted on two widely used benchmarks, ScienceQA~\citep{lu2022learn_sqa} and SEED-Bench-IMG~\citep{li2024seed}, consistent with prior studies on VLMs such as LLaVA, using VLMEvalKit~\citep{duan2024vlmevalkit} tool.
For comparison, WSVD is benchmarked against several baselines, including SVD-based approaches (ASVD~\citep{yuan2023asvd}, SVD-LLM~\citep{wang2024svd}, QSVD~\citep{wang2025qsvd}) and quantization-based techniques (DuQuant~\citep{lin2024duquant}, QVLM~\citep{wang2024qvlm}). For ASVD, SVD-LLM and QSVD, we follow their official implementations and apply SVD independently to the $Q,K,V$ matrices to ensure a fair comparison with WSVD, while leaving other linear layers unchanged. More results are shown in the Appendix. %DuQuant and QVLM are applied directly according to their public repositories. All methods use the same calibration set and random seed for fairness.

To isolate the impact of SVD from quantization, we introduce \textbf{WSVD-noQ} (Section~\ref{sec:wsvd}), which applies only the SVD techniques described in Sections~\ref{sec:per-head-svd} and~\ref{sec:wsvd}. We compare it with ASVD, SVD-LLM, and QSVD-noQ (unquantized version of QSVD).
We then apply QAT in Section~\ref{sec:qat} on top of WSVD-noQ, benchmarking against DuQuant, QVLM, and QSVD. We also evaluate QASVD, which applies QuaRot~\citep{ashkboos2024quarot} to the SVD-truncated VLMs produced by ASVD. For fine-tuning and QAT, we adopt lightweight local optimization to minimize overhead. $A_h$ and $B_h$ are updated with Adam~\citep{kingma2014adam} (learning rate $1\times10^{-4}$ for fine-tuning and $1\times10^{-5}$ for QAT), while $S_2$ is updated during QAT using the Cayley optimizer~\citep{wen2013feasible}. Local fine-tuning is performed for 100 steps and QAT for 50 steps, ensuring effective adaptation while keeping the additional latency negligible.
All experiments are conducted on NVIDIA H100 GPUs.

%Results are reported in Section~\ref{sec:eval-noq} and Section~\ref{sec:eval-q}. For fine-tuning and QAT, we adopt lightweight local optimization to minimize overhead. $A_h$ and $B_h$ are updated with Adam~\citep{kingma2014adam} (learning rate $1\times10^{-4}$ for fine-tuning and $1\times10^{-5}$ for QAT), while $S_2$ is updated during QAT using the Cayley optimizer~\citep{wen2013feasible} (learning rate $1$). Local fine-tuning is performed for 100 steps and QAT for 50 steps, ensuring effective adaptation while keeping the additional latency negligible. All performance experiments are conducted on NVIDIA H100 GPUs. Ablation studies are presented in Section~\ref{sec:eval-ablation}, and system-level evaluation, including decoding stage latency, is reported in Section~\ref{sec:eval-system}.   

\subsection{Accuracy Evaluation on WSVD-noQ}
\label{sec:eval-noq}
We first evaluate the FP16 performance of WSVD-noQ under different rank budgets. To ensure fairness, we align the parameter ratio $\rho_1$ across all methods. For WSVD, $\rho_1$ is defined in Equation~\ref{eq:per-head-orig}, while for other SVD-based baselines, $\rho_1$ is defined as the proportion of parameters relative to the original model after SVD is applied. 

\begin{table*}[t]
    \newcommand{\bc}{\cellcolor{blue!10}}
    \newcommand{\accval}[1]{\fontsize{9.8pt}{10pt}\selectfont #1}
    \newcommand{\accvalours}[1]{\fontsize{9.75pt}{10pt}\selectfont \textbf{#1}}
    \centering
    \small
    \caption{Accuracy evaluation of different methods under FP16 (detailed results in Appendix~\ref{sec:noq_appx}).}
    \vspace{-5pt}
    \resizebox{\linewidth}{!}{ %
    \begin{tabular}{l|l|c|c|c|c|c|c|c|c|c|c|c}
    \toprule
    
    \multirow{2}{*}{Acc.} & \multirow{2}{*}{Method} & \multicolumn{5}{c|}{ScienceQA-IMG $\uparrow$} & \multicolumn{5}{c|}{SEED-Bench $\uparrow$} & \multirow{2}{*}{Avg. $\uparrow$ } \\
    \cmidrule{3-12}
    & & $\rho_1: 90\%$ & $\rho_1: 80\%$ & $\rho_1: 70\%$ & $\rho_1: 60\%$ & $\rho_1: 50\%$ 
    & $\rho_1: 90\%$ & $\rho_1: 80\%$ & $\rho_1: 70\%$ & $\rho_1: 60\%$ & $\rho_1: 50\%$ \\
    
    \midrule
    
    \multirow{5.75}{*}{\rotatebox{90}{\makecell{LLaVA-v1.5\\7B}}}   

    & ASVD  
        & \accval{49.93\%} & \accval{50.12\%} & \accval{47.10\%} & \accval{36.69\%} & \accval{19.19\%} 
        & \accval{54.27\%} & \accval{53.53\%} & \accval{48.35\%} & \accval{37.17\%} & \accval{24.17\%} 
        & \accval{42.05\%}\\
    & SVD-LLM   
        & \accval{65.44\%} & \accval{63.71\%} & \accval{61.92\%} & \accval{57.41\%} & \accval{55.53\%}
        & \accval{57.89\%} & \accval{57.50\%} & \accval{55.33\%} & \accval{54.64\%} & \accval{55.31\%} 
        & \accval{58.47\%}\\
    & QSVD-noQ   
        & \accval{67.72\%} & \accval{\textbf{68.22\%}} & \accval{67.08\%} & \accval{65.05\%} & \accval{62.37\%} 
        & \accval{59.84\%} & \accval{59.07\%} & \accval{59.78\%} & \accval{59.00\%} & \accval{58.23\%} 
        & \accval{62.64\%}\\
    % & \rowcolor{blue!10}\textbf{WSVD-noQ} 
    & \bc\textbf{WSVD-noQ}
        & \bc\accval{\textbf{68.17\%}} & \bc\accval{67.72\%} & \bc\accval{\textbf{67.28\%}} & \bc\accval{\textbf{65.89\%}} & \bc\accval{\textbf{65.49\%}} 
        & \bc\accval{\textbf{60.10\%}} & \bc\accval{\textbf{60.17\%}} & \bc\accval{\textbf{59.89\%}} & \bc\accval{\textbf{60.18\%}} & \bc\accval{\textbf{60.46\%}} 
        & \bc\accval{\textbf{63.54\%}}\\
    \cline{3-12}
    \rule{0pt}{2.75ex}
    & \textcolor{gray}{FP16} & \multicolumn{5}{c|}{\textcolor{gray}{Accuracy: 68.01\%}} & \multicolumn{5}{c|}{\textcolor{gray}{Accuracy: 60.18\%}} & \textcolor{gray}{\accval{64.10\%}}  \\
    
    \midrule
    
    \multirow{5.75}{*}{\rotatebox{90}{\makecell{LLaVA-Next\\13B}}} %LLaVA-v1.5-13B

    & ASVD   
        & \accval{71.24\%} & \accval{70.60\%} & \accval{71.44\%} & \accval{71.38\%} & \accval{69.81\%} 
        & \accval{70.88\%} & \accval{70.26\%} & \accval{70.01\%} & \accval{69.69\%} & \accval{69.01\%} 
        & \accval{70.43\%}\\
    & SVD-LLM   
        & \accval{72.53\%} & \accval{72.24\%} & \accval{71.74\%} & \accval{71.15\%} & \accval{70.55\%} 
        & \accval{70.76\%} & \accval{70.63\%} & \accval{70.25\%} & \accval{69.96\%} & \accval{69.58\%} 
        & \accval{70.94\%}\\
    & QSVD-noQ   
        & \accval{71.94\%} & \accval{72.14\%} & \accval{71.74\%} & \accval{72.14\%} & \accval{71.79\%} 
        & \accval{71.23\%} & \accval{71.02\%} & \accval{71.06\%} & \accval{70.92\%} & \accval{70.40\%} 
        & \accval{71.44\%}\\
    & \bc\textbf{WSVD-noQ} 
        & \bc\accval{\textbf{72.88\%}} & \bc\accval{\textbf{72.98\%}} & \bc\accval{\textbf{73.57\%}} & \bc\accval{\textbf{73.48\%}} & \bc\accval{\textbf{73.28\%}} 
        & \bc\accval{\textbf{71.29\%}} & \bc\accval{\textbf{71.17\%}} & \bc\accval{\textbf{71.25\%}} & \bc\accval{\textbf{70.95\%}} & \bc\accval{\textbf{70.81\%}} 
        & \bc\accval{\textbf{72.17\%}}\\
    \cline{3-12}
    \rule{0pt}{2.75ex}
    & \textcolor{gray}{FP16} & \multicolumn{5}{c|}{\textcolor{gray}{Accuracy: 73.23\%}} & \multicolumn{5}{c|}{\textcolor{gray}{Accuracy: 71.30\%}} & \textcolor{gray}{\accval{72.27\%}}\\

    \midrule
    
    \multirow{7}{*}{\rotatebox{90}{\makecell{SmolVLM\\2B}}}
    %\cline{2-14}
    %\rule{0pt}{2.25ex}
    %& & $\rho_1: 90\%$ & $\rho_1: 80\%$ & $\rho_1: 70\%$ & & & $\rho_1: 90\%$ & $\rho_1: 80\%$ & $\rho_1: 70\%$ & & \\
    
    & &
    \multicolumn{5}{l|}{%
      \begin{tabularx}{0.575\textwidth}{Y|Y|Y}
        $\rho_1: 90\%$ & $\rho_1: 80\%$ & $\rho_1: 70\%$
      \end{tabularx}
    }
    &
    \multicolumn{5}{l|}{%
      \begin{tabularx}{0.575\textwidth}{Y|Y|Y}
        $\rho_1: 90\%$ & $\rho_1: 80\%$ & $\rho_1: 70\%$
      \end{tabularx}
    } & \\    
    \cline{2-13}
    \rule{0pt}{2.75ex}
    & ASVD  
        & 
        \multicolumn{5}{l|}{%
          \begin{tabularx}{0.575\textwidth}{Y|Y|Y}
            \accval{29.30\%} & \accval{3.97\%} & \accval{0.20\%}
          \end{tabularx}
        }
        & \multicolumn{5}{l|}{%
          \begin{tabularx}{0.575\textwidth}{Y|Y|Y}
            \accval{17.85\%} & \accval{1.50\%} & \accval{0.95\%}
          \end{tabularx}
        }
        & \accval{8.96\%} \\
    & SVD-LLM   
        & 
        \multicolumn{5}{l|}{%
          \begin{tabularx}{0.575\textwidth}{Y|Y|Y}
            \accval{40.06\%} & \accval{17.20\%} & \accval{3.82\%}
          \end{tabularx}
        }
        & 
        \multicolumn{5}{l|}{%
          \begin{tabularx}{0.575\textwidth}{Y|Y|Y}
            \accval{32.49\%} & \accval{15.89\%} & \accval{4.60\%}
          \end{tabularx}
        }
        & \accval{19.01\%}\\
    & QSVD-noQ   
        & 
        \multicolumn{5}{l|}{%
          \begin{tabularx}{0.575\textwidth}{Y|Y|Y}
            \accval{\textbf{77.00\%}} & \accval{62.77\%} & \accval{42.59\%}
          \end{tabularx}
        }
        & 
        \multicolumn{5}{l|}{%
          \begin{tabularx}{0.575\textwidth}{Y|Y|Y}
            \accval{64.80\%} & \accval{50.46\%} & \accval{36.24\%}
          \end{tabularx}
        }
        & \accval{55.64\%}\\
    & \bc\textbf{WSVD-noQ} 
        & 
        \multicolumn{5}{l|}{%
          \bc\begin{tabularx}{0.575\textwidth}{Y|Y|Y}
            \accval{76.30\%} & \accval{\textbf{71.74\%}} & \accval{\textbf{60.93\%}}
          \end{tabularx}
        }
        & 
        \multicolumn{5}{l|}{%
          \bc\begin{tabularx}{0.575\textwidth}{Y|Y|Y}
            \accval{\textbf{65.78\%}} & \accval{\textbf{63.29\%}} & \accval{\textbf{54.45\%}}
          \end{tabularx}
        }
        & \bc\accval{\textbf{65.42\%}}\\
    \cline{3-12}
    \rule{0pt}{2.75ex}
    & \textcolor{gray}{FP16} & \multicolumn{5}{c|}{\textcolor{gray}{Accuracy: 84.53\%}} & \multicolumn{5}{c|}{\textcolor{gray}{Accuracy: 68.47\%}} & \textcolor{gray}{\accval{76.53\%}} \\
    
    \bottomrule
    \end{tabular}
    }
    \label{tab:svd-result}
    \vspace{-10pt}
\end{table*}
The evaluation results are summarized in Table~\ref{tab:svd-result} (details in Appendix~\ref{sec:noq_appx}). Under the same parameter ratio $\rho_1$, WSVD-noQ surpasses ASVD, SVD-LLM, and QSVD-noQ in accuracy in most cases. On large-scale models such as LLaVA-v1.5 13B and LLaVA-Next 13B, WSVD-noQ incurs less than a $1\%$ accuracy drop on ScienceQA-IMG and SEED-Bench compared to the FP16 baseline. Notably, for LLaVA-Next 13B, when $\rho_1 \leq 70\%$, WSVD-noQ even outperforms the FP16 model on ScienceQA-IMG. For example, at $\rho_1 = 70\%$, WSVD-noQ reaches $73.57\%$ accuracy, exceeding the FP16 baseline by more than $0.3\%$. This suggests that low-rank approximation may implicitly mitigate hallucinations~\citep{liu2024survey}, though further validation is required.
Furthermore, WSVD-noQ delivers consistently higher average accuracy across datasets and parameter ratios. The advantage over other baselines becomes increasingly evident as $\rho_1$ decreases. For example, on SmolVLM, WSVD-noQ attains over $60\%$ accuracy on ScienceQA-IMG, while baselines fail to yield usable results under the same parameter ratio settings.

\subsection{Accuracy Evaluation of WSVD}
\label{sec:eval-q}
We present results under two weight–activation quantization configurations: W8A8 for WSVD with rank settings $\rho_{1}=50\%$ and $\rho_{2}\approx50\%$, and W8A4 for all other baselines. This design keeps cache size and parameter size comparable across methods, while WSVD’s rank truncation further reduces its parameter budget, ensuring fairness in comparison. 

%For QASVD and QSVD, we adopt $\rho_1 = 100\%$ truncation to match the W8 parameter budget of other baselines. Since these methods do not employ per-head SVD, they cannot leverage latent cache representations to accelerate inference and must store the full KV cache. As a result, they are also evaluated under the A4 setting to ensure comparable cache sizes. 

For activation quantization, we adopt per-token symmetric quantization. For weight quantization, we employ round-to-nearest (RTN) with per-channel symmetric scaling and a learnable clipping ratio, where the clipping value is selected via linear search to minimize squared error, following QuaRot~\citep{ashkboos2024quarot}. This quantization scheme is applied to the per-head Q/K/V weight matrices and all remaining attention and feed-forward modules, ensuring that the dominant matrix multiplications in each transformer block are executed in low precision. As shown in Table~\ref{tab:svd-qat-result}, WSVD consistently outperforms the baselines in most cases, despite using a smaller parameter budget and the same cache size. On average across models and datasets, WSVD incurs only a modest accuracy drop of just over 1\% relative to the FP16 baseline, while reducing cache size to 25\% of the FP16 model. At the same time, WSVD achieves more than 1\% higher average accuracy than all baselines, demonstrating that the integration of per-head SVD and quantization with WSVD only lead to minimized accuracy loss.

%not only compresses model and cache size but also helps alleviate the I/O bottleneck, thereby enabling further decoding acceleration.

\begin{table*}[t]
    \newcommand{\bc}{\cellcolor{blue!10}}
    \newcommand{\accval}[1]{\fontsize{9.8pt}{10pt}\selectfont #1}
    \newcommand{\accvalours}[1]{\fontsize{9.75pt}{10pt}\selectfont \textbf{#1}}
    \centering
    \small
    \caption{Accuracy evaluation of different methods under low-precision on LLaVA-v1.5 7B, LLaVA-v1.5 13B, LLaVA-Next 7B and LLaVA-Next 13B.}
    \vspace{-5pt}
    \resizebox{0.9\linewidth}{!}{ %
    \begin{tabular}{l|c|c|c|c|c|c|c|c|c}
    \toprule
    
    \multirow{2}{*}{Method} & \multicolumn{4}{c|}{ScienceQA-IMG $\uparrow$} & \multicolumn{4}{c|}{SEED-Bench $\uparrow$} & \multirow{2}{*}{Avg. $\uparrow$ } \\
    \cmidrule{2-9}
    & v1.5 7B & v1.5 13B & Next 7B & Next 13B & v1.5 7B & v1.5 13B & Next 7B & Next 13B & \\
    
    \midrule
    
    % QuaRot  
    %     & \accval{63.19\%} & \accval{68.02\%} & \accval{64.53\%} & \accval{66.98\%} & \accval{58.18\%} & \accval{58.53\%} & \accval{69.09\%} & \accval{71.15\%} & \accval{64.96\%} \\
    DuQuant   
        & \accval{57.36\%} & \accval{67.22\%} & \accval{66.34\%} & \accval{70.20\%} & \accval{54.11\%} & \accval{61.43\%} & \accval{63.64\%} & \accval{66.15\%} & \accval{63.31\%} \\
    QVLM 
        & \accval{55.24\%} & \accval{66.46\%} & \accval{60.60\%} & \accval{65.28\%} & \accval{50.13\%} & \accval{59.22\%} & \accval{50.38\%} & \accval{65.39\%} & \accval{59.09\%} \\
    QASVD 
        & \accval{41.92\%} & \accval{65.34\%} & \accval{49.37\%} & \accval{64.85\%} & \accval{41.26\%} & \accval{59.30\%} & \accval{49.63\%} & \accval{66.54\%} & \accval{54.78\%} \\
    QSVD 
        & \accval{\textbf{65.61\%}} & \accval{70.12\%} & \accval{66.10\%} & \accval{70.43\%} & \accval{58.49\%} & \accval{\textbf{62.95\%}} & \accval{65.63\%} & \accval{69.21\%} & \accval{66.07\%} \\
    \rowcolor{blue!10}\textbf{WSVD} 
        & \accval{64.25\%} & \accval{\textbf{72.14\%}} & \accval{\textbf{66.94\%}} & \accval{\textbf{73.08\%}} & \accval{\textbf{60.23\%}} & \accval{62.01\%} & \accval{\textbf{67.49\%}} & \accval{\textbf{70.67\%}} & \accval{\textbf{67.10\%}} \\
    \cline{2-10}
    \rule{0pt}{2.75ex}
    \textcolor{gray}{FP16} & 
    \textcolor{gray}{\accval{68.10\%}} &         \textcolor{gray}{\accval{71.83\%}} & 
    \textcolor{gray}{\accval{69.60\%}} & 
    \textcolor{gray}{\accval{73.23\%}} & 
    \textcolor{gray}{\accval{60.18\%}} & 
    \textcolor{gray}{\accval{62.54\%}} & 
    \textcolor{gray}{\accval{69.02\%}} & 
    \textcolor{gray}{\accval{71.30\%}} & 
    \textcolor{gray}{\accval{68.23\%}} \\
    
    \bottomrule
    \end{tabular}
    }
    \label{tab:svd-qat-result}
    %\vspace{-10pt}
\end{table*}   

\subsection{Ablation Study}
\label{sec:eval-ablation}

\paragraph{Effectiveness of Weighted Local Finetuning}
We evaluate the impact of WSVD fine-tuning, as described in Section~\ref{sec:wsvd}, on accuracy performance using ScienceQA-IMG for WSVD-noQ. The comparison is made against the WSVD-noQ baseline, which applies standard SVD without accounting for the relative importance of weight elements, while keeping all other settings identical.
As shown in Table~\ref{tab:scienceqa-wsvd-ablation}, WSVD-noQ consistently outperforms the no-finetuning variant (WSVD-noFT), demonstrating that incorporating relative weight importance during the SVD process leads to significantly improved performance over standard SVD.

%under tighter memory and computational budgets (e.g., $\rho_{1}{=}70\%$ and $\rho_{1}{=}50\%$), the introduction of local finetuning achieves up to \textbf{+1.4\%} improvement over the baseline. 
%These results validate the effectiveness of WSVD fine-tuning in enhancing model robustness under compression. 

% \begin{wraptable}{r}{0.45\textwidth}
%   \centering
%   \small
%     \caption{Results of weighted finetuning ablation.}

%   \resizebox{1\linewidth}{!}{
%     \begin{tabular}{l|c|c|c|c}
%       \toprule
%       Acc. & Method & $\rho_{1}{=}90\%$ & $\rho_{1}{=}70\%$ & $\rho_{1}{=}50\%$ \\
%       \midrule
%       \multirow{4}{*}{\rotatebox{90}{\makecell{v1.5 \\13B}}}
%       & \textcolor{gray}{FP16} & \multicolumn{3}{c}{\textcolor{gray}{71.83\%}}\\
%       \cline{2-5}
%       \rule{0pt}{2.75ex}
%       & W/o WSVD FT & 71.59\% & 72.38\% & \textbf{71.49\%} \\
%       & WSVD-noQ & \textbf{71.99\%} & \textbf{72.53\%} & 71.44\% \\
%       \midrule
%       \multirow{4}{*}{\rotatebox{90}{\makecell{Next \\13B}}}
%       & \textcolor{gray}{FP16} & \multicolumn{3}{c}{\textcolor{gray}{73.23\%}}\\
%       \cline{2-5}
%       \rule{0pt}{2.75ex}
%       & W/o WSVD FT & \textbf{72.93\%} & 72.93\% & 73.08\% \\
%       & WSVD-noQ & 72.88\% & \textbf{73.57\%} & \textbf{73.28\%} \\
%       \bottomrule
%     \end{tabular}
% }
%   \label{tab:scienceqa-wsvd-ablation}
% \end{wraptable}

\begin{table}[t]
  \centering
  \small
    \caption{Results of weighted finetuning ablation under different $\rho_{1}$ settings.}
    \vspace{-5pt}
  \resizebox{1\linewidth}{!}{
    \begin{tabular}{l|c|c|c|c}
      \toprule
      Acc. & Method & $\rho_{1}{=}90\%$ & $\rho_{1}{=}70\%$ & $\rho_{1}{=}50\%$ \\
      \midrule
      \multirow{4}{*}{\rotatebox{90}{\makecell{v1.5 \\7B}}}
      & \textcolor{gray}{FP16} & \multicolumn{3}{c}{\textcolor{gray}{68.01\%}}\\
      \cline{2-5}
      \rule{0pt}{2.75ex}
      & WSVD-noFT & 67.82\% & 66.82\% & 65.09\% \\
      & WSVD-noQ & \textbf{68.17\%} & \textbf{67.28\%} & \textbf{65.49\%} \\
      \bottomrule
    \end{tabular}
    \hspace{1cm}
      \begin{tabular}{l|c|c|c|c}
      \toprule
      Acc. & Method & $\rho_{1}{=}90\%$ & $\rho_{1}{=}70\%$ & $\rho_{1}{=}50\%$ \\
      \midrule
      \multirow{4}{*}{\rotatebox{90}{\makecell{Next \\7B}}}
      & \textcolor{gray}{FP16} & \multicolumn{3}{c}{\textcolor{gray}{69.60\%}}\\
      \cline{2-5}
      \rule{0pt}{2.75ex}
      & WSVD-noFT & 69.76\% & 68.61\% & 66.46\% \\
      & WSVD-noQ & \textbf{69.81\%} & \textbf{69.36\%} & \textbf{67.87\%} \\
      \bottomrule
    \end{tabular}
}
  \label{tab:scienceqa-wsvd-ablation}
  \vspace{-10pt}
\end{table}

\begin{wraptable}{r}{0.5\textwidth}
\newcommand{\accval}[1]{\fontsize{9.8pt}{10pt}\selectfont #1}
  \vspace{-15pt}
  \centering
  \small
    \caption{Results of local QAT ablation.}
    \vspace{-5pt}
    \resizebox{\linewidth}{!}{ %
    \begin{tabular}{c|c|c|c|c|c}
    \toprule
    
    \multirow{2}{*}{Method} & \multicolumn{4}{c|}{ScienceQA-IMG $\uparrow$} & \multirow{2}{*}{Avg. $\uparrow$ } \\
    \cmidrule{2-5}
    & v1.5 7B & v1.5 13B & Next 7B & Next 13B \\
    
    \midrule
    W/o QAT  
        & \accval{63.91\%} & \accval{71.99\%} & \accval{66.59\%} & \accval{72.68\%} & \accval{68.79\%}  \\
    WSVD 
        & \accval{\textbf{64.25\%}} & \accval{\textbf{72.14\%}} & \accval{\textbf{66.94\%}} & \accval{\textbf{73.08\%}} & \accval{\textbf{69.10\%}}  \\
    \bottomrule
    \end{tabular}
}
  \label{tab:wsvd-qat-ablation}
  \vspace{-5pt}
\end{wraptable}

\paragraph{Effectiveness of QAT}
We further examine the impact of local QAT on the low-rank model, as described in Section~\ref{sec:qat}. Specifically, we compare WSVD against a baseline that uses the same quantization settings but does not fine-tune $S_{2}$, $A_{h}$, or $B_{h}$ mentioned in Section~\ref{sec:qat}, while keeping all other settings identical. As shown in Table~\ref{tab:wsvd-qat-ablation}, under W8A8, WSVD consistently surpasses the baseline across all models. These results demonstrate that local QAT effectively recovers the performance lost due to low-precision quantization.

%Finally, we study the role of local QAT (Section~\ref{sec:qat}) on ScienceQA-IMG under the same settings as the WSVD quantization experiments. For the “w/o QAT” variant, models are directly quantized after local fine-tuning without QAT. 
%As shown in Table~\ref{tab:wsvd-qat-ablation}, WSVD consistently outperforms the baseline across all models, yielding an average improvement of about $0.3\%$. This demonstrates that local QAT effectively recovers performance lost to low-precision quantization.

\subsection{System Evaluation on VLM}
\label{sec:eval-system}
\begin{wrapfigure}{r}{0.5\linewidth}
    \vspace{-40pt}
    \centering
    \includegraphics[width=\linewidth]{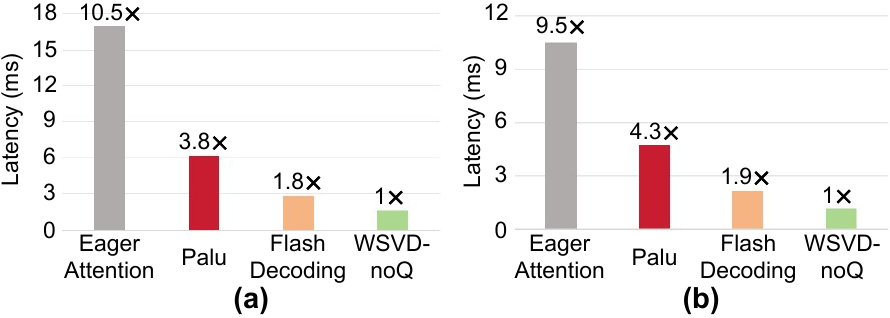}
    \vspace{-20pt}
    \caption{Latency evaluation and normalized latency on: (a) RTX 4090 and (b) RTX 5090.}
    \vspace{-10pt}
    \label{fig:eval-latency}
\end{wrapfigure}
We assess the system-level performance of WSVD-noQ, with a focus on decoding-stage acceleration. Specifically, we measure the layer-wise decoding latency of LLaVA-Next 7B across the attention and feed-forward modules using our fused kernel implementation described in Section~\ref{sec:wsvd-sys} on RTX 4090 and 5090 GPUs.
For comparison, we include Eager Attention without Flash Decoding, Palu~\citep{chang2024palu}, and Flash Decoding~\citep{dao2023flashdecoding} as the baseline algorithms. 
For Flash Decoding, we adopt scaled dot-product attention (SDPA), while Palu is evaluated using its official repository.
Both Eager Attention and Flash Decoding operate on the full KV cache, while Palu and WSVD-noQ restrict the latent size to $\rho_{2}=50\%$, corresponding to $\rho_{1}\approx51.5\%$ for WSVD.
All measurements are conducted with a batch size of $16$ and a sequence length of $8192$. Since Palu supports only batch size $1$, we use an equivalent sequence length of $16 \times 8192$ for fair comparison. In addition, we report latency results of full-matrix SVD and per-head SVD in Appendix~\ref{sec:latency-appx}.

As shown in Figure~\ref{fig:eval-latency}, WSVD-noQ consistently outperforms all baselines on both GPUs in latency. Relative to Flash Decoding, it achieves over $1.8\times$ speedup, driven by reduced I/O overhead and negligible reconstruction cost enabled by our scheme. Compared with Palu, WSVD-noQ attains lower latency through two advantages: algorithmically, per-head SVD provides finer-grained compression than Palu’s group-head SVD; system-wise, our fused kernel is fully integrated into the flash decoding pipeline. These results demonstrate that WSVD, together with our fused kernel design, offers an effective system-level solution that alleviates I/O bottlenecks and enables practical decoding acceleration in VLMs while maintaining accuracy performance as the original model.

\begin{table}[h]
    \centering
    \caption{Latency (ms) on RTX 4090 (left) and RTX 5090 (right) for different sequence lengths.}
    \vspace{-5pt}
    \label{tab:seqlen_latency_4090_5090}

    % 4090
    \begin{subtable}[t]{0.48\linewidth}
        \centering
        \resizebox{\linewidth}{!}{%
        \begin{tabular}{lcccccc}
            \toprule
            Seq Len      & 1024 & 2048 & 4096 & 8192 & 16K  & 32K  \\
            \midrule
            Flash Decoding & 0.92 & 1.21 & 1.77 & 2.92 & 5.14 & 9.64 \\
            WSVD-noQ       & 0.70 & 0.83 & 1.12 & 1.66 & 2.89 & 5.28 \\
            Speedup        & 1.3$\times$ & 1.5$\times$ & 1.6$\times$ & 1.8$\times$ & 1.8$\times$ & 1.8$\times$ \\
            \bottomrule
        \end{tabular}
        }%
    \end{subtable}\hfill
    %
    % 5090
    \begin{subtable}[t]{0.48\linewidth}
        \centering
        \resizebox{\linewidth}{!}{%
        \begin{tabular}{lcccccc}
            \toprule
            Seq Len      & 1024 & 2048 & 4096 & 8192 & 16K  & 32K  \\
            \midrule
            Flash Decoding & 0.65 & 0.86 & 1.28 & 2.14 & 3.81 & 7.18 \\
            WSVD-noQ       & 0.58 & 0.66 & 0.83 & 1.15 & 1.79 & 3.06 \\
            Speedup        & 1.1$\times$ & 1.3$\times$ & 1.5$\times$ & 1.9$\times$ & 2.1$\times$ & 2.3$\times$ \\
            \bottomrule
        \end{tabular}
        }%
    \end{subtable}
    \vspace{-5pt}
\end{table}

\begin{table}[h]
    \centering
    \caption{Latency (ms) on RTX 4090 (left) and RTX 5090 (right) for different batch sizes.}
    \vspace{-5pt}
    \label{tab:bsz_latency_4090_5090}

    % 4090
    \begin{subtable}[t]{0.42\linewidth}
        \centering
        \resizebox{\linewidth}{!}{%
        \begin{tabular}{lccccc}
            \toprule
            Batch Size           & 4    & 8    & 16   & 32   & 64   \\
            \midrule
            Flash Decoding & 1.13 & 1.71 & 2.92 & 5.11 & 9.67 \\
            WSVD-noQ       & 0.82 & 1.11 & 1.66 & 2.91 & 5.33 \\
            Speedup        & 1.4$\times$ & 1.5$\times$ & 1.8$\times$ & 1.8$\times$ & 1.8$\times$ \\
            \bottomrule
        \end{tabular}
        }%
    \end{subtable}
    \hspace{0.05\linewidth}
    %
    % 5090
    \begin{subtable}[t]{0.42\linewidth}
        \centering
        \resizebox{\linewidth}{!}{%
        \begin{tabular}{lccccc}
            \toprule
            Batch Size           & 4    & 8    & 16   & 32   & 64   \\
            \midrule
            Flash Decoding & 0.75 & 1.07 & 2.14 & 3.38 & 5.97 \\
            WSVD-noQ       & 0.66 & 0.84 & 1.15 & 1.79 & 3.11 \\
            Speedup        & 1.1$\times$ & 1.3$\times$ & 1.9$\times$ & 1.9$\times$ & 1.9$\times$ \\
            \bottomrule
        \end{tabular}
        }%
    \end{subtable}
    \vspace{-10pt}
\end{table}

\paragraph{Impact of Sequence Length and Batch Size}
We further perform an ablation over various sequence lengths (Table~\ref{tab:seqlen_latency_4090_5090}) and batch sizes (Table~\ref{tab:bsz_latency_4090_5090}) for LLaVA-Next 7B under the same setting, and report the layer-wise decoding latency of Flash Decoding and WSVD-noQ on RTX 4090 and RTX 5090 GPUs. With batch size 16, as the sequence length grows from 1K to 32K tokens, WSVD-noQ improves over Flash Decoding by about $1.3\times$ to $1.8\times$ on RTX 4090 and up to $2.35\times$ on RTX 5090. For a fixed 8192 sequence context, increasing batch size from 4 to 64 yields roughly $1.4\times$ to $1.9\times$ speedups on both GPUs. This trend reflects that longer contexts make KV-cache I/O increasingly dominant, so our WSVD-based compression and decoding kernel delivers larger relative gains.

\paragraph{Impact of Rank Ratio}
Using the same setting as Section~\ref{sec:eval-system}, we vary the rank ratio $\rho_2 \in \{90\%, 70\%, 50\%\}$ for WSVD-noQ and measure the latency on RTX~4090 and RTX~5090 GPUs. Table~\ref{tab:rank-latency} summarizes the impact of rank ratio on decoding latency. Smaller $\rho_2$ values (i.e., lower ranks) consistently yield lower latency, demonstrating that WSVD’s fused kernel can effectively translate rank reduction into tangible decoding speedups over the Flash Decoding baseline.

% \begin{table}[h]
%     \centering
%     \small
%     \caption{Decoding latency (ms) under different $\rho_2$.}
%     \begin{tabular}{lcccc}
%         \toprule
%         GPU & Flash Decoding & $\rho_2:90\%$ & $\rho_2:70\%$ & $\rho_2:50\%$ \\
%         \midrule
%         4090 & 2.92 & 2.83 & 2.53 & 1.66 \\
%         5090 & 2.14 & 1.87 & 1.75 & 1.15 \\
%         \bottomrule
%     \end{tabular}
%     \label{tab:rank-latency}
% \end{table}

% \begin{table}[h]
%     \centering
%     \caption{Latency (ms) on RTX 3060 for different sequence lengths (bsz $=16$).}
%     \label{tab:seqlen_latency_3060}
%     \resizebox{0.4\linewidth}{!}{%
%     \begin{tabular}{lccccc}
%         \toprule
%         Seq Len        & 1024 & 2048 & 4096 & 8192 & 16K  \\
%         \midrule
%         Flash Decoding & 3.37 & 4.88 & 7.81 & 13.27 & 24.59 \\
%         WSVD-noQ       & 2.18 & 2.68 & 3.62 & 5.54 & 9.49 \\
%         Speedup        & 1.5$\times$ & 1.8$\times$ & 2.2$\times$ & 2.4$\times$ & 2.6$\times$ \\
%         \bottomrule
%     \end{tabular}
%     }%
% \end{table}
\begin{table}[h]
    \centering
    \small
    \begin{minipage}{0.48\linewidth}
        \centering
        \captionof{table}{Latency (ms) under different $\rho_2$.}
        \vspace{-5pt}
        \label{tab:rank-latency}
        \resizebox{\linewidth}{!}{%
        \begin{tabular}{lcccc}
            \toprule
            GPU & Flash Dec. & $\rho_2{:}90\%$ & $\rho_2{:}70\%$ & $\rho_2{:}50\%$ \\
            \midrule
            4090 & 2.92 & 2.83 & 2.53 & 1.66 \\
            5090 & 2.14 & 1.87 & 1.75 & 1.15 \\
            \bottomrule
        \end{tabular}
        }%
    \end{minipage}
    \hfill
    \begin{minipage}{0.48\linewidth}
        \centering
        \captionof{table}{Latency (ms) on RTX 3060 ($\rho_2:50\%$).}
        \vspace{-5pt}
        \label{tab:seqlen_latency_3060}
        \resizebox{\linewidth}{!}{%
        \begin{tabular}{lccccc}
            \toprule
            Seq Len        & 1024 & 2048 & 4096 & 8192 & 16K  \\
            \midrule
            Flash Decoding & 3.37 & 4.88 & 7.81 & 13.27 & 24.59 \\
            WSVD-noQ       & 2.18 & 2.68 & 3.62 & 5.54 & 9.49 \\
            Speedup        & 1.5$\times$ & 1.8$\times$ & 2.2$\times$ & 2.4$\times$ & 2.6$\times$ \\
            \bottomrule
        \end{tabular}
        }%
    \end{minipage}
    \vspace{-10pt}
\end{table}

\paragraph{Speedup on Low-end GPU}
To evaluate our method on more modest hardware, we benchmark the latency of LLaVA-Next 7B with WSVD-noQ on an RTX 3060 (Table \ref{tab:seqlen_latency_3060}) under the same setting, and compare it with the Flash Decoding baseline. On RTX 3060, WSVD-noQ reduces latency from 3.37\,ms to 2.18\,ms at 1K tokens ($1.55\times$) and from 24.59\,ms to 9.49\,ms at 16K tokens ($2.59\times$). These speedups are larger than on 4090/5090-class GPUs because the lower memory bandwidth of RTX 3060 makes KV-cache I/O more dominant. By shrinking the KV cache and using a fused decoding kernel, WSVD reduces memory traffic and achieves larger latency gains on low-end devices.

%As shown in Figure~\ref{fig:eval-latency}, WSVD-noQ consistently outperforms all baselines on both GPUs. Relative to Flash Decoding, it achieves up to $1.8\times$ speedup, primarily due to reduced I/O overhead and negligible reconstruction cost enabled by per-head SVD and the latent cache. Compared to Palu, WSVD-noQ delivers lower latency by combining two advantages: (i) algorithmically, per-head SVD yields finer-grained compression than Palu’s group-head SVD; and (ii) system-wise, our fused kernel is fully integrated into the flash decoding pipeline. These results show that WSVD, combined with our fused kernel design, provides an effective system-level solution that alleviates I/O bottlenecks and enables practical decoding acceleration in VLMs, while preserving performance under compression.

\section{Conclusion}
In this work, we present WSVD, a weighted low-rank approximation framework that integrates per-head SVD, weighted fine-tuning, and quantization-aware training to compress and accelerate vision–language models. By aligning algorithmic design with system-level optimization through our fused kernel, WSVD achieves over $1.8\times$ decoding speedup while preserving accuracy under aggressive compression. %Our results demonstrate that fine-grained decomposition and importance-aware adaptation are key to overcoming the latency and memory bottlenecks of prior SVD-based methods. Looking ahead, we believe the principles behind WSVD can be extended to broader multimodal architectures and hardware platforms, further advancing efficient large-model deployment.

\section*{Ethics Statement}
This work focuses on model compression and acceleration techniques for vision–language models. 
All datasets used in this study (ScienceQA-IMG and SEED-Bench) are publicly available and widely adopted in the community. 
Our research does not involve human subjects, private or sensitive data, or personally identifiable information. 
The proposed method aims to improve the efficiency of large models, which may contribute to reducing the computational and environmental costs of deployment. 
We are not aware of any direct ethical concerns specific to this work.

\section*{Reproducibility Statement}
We take reproducibility seriously and provide the following details: 
\begin{itemize}
    \item \textbf{Code and models:} We will release the full implementation of WSVD, including training and inference code, as well as evaluation scripts, upon publication.  
    \item \textbf{Datasets:} All datasets used in this work are publicly available. In particular, we evaluate on ScienceQA-IMG and SEED-Bench, both of which can be accessed without restriction. We will also provide preprocessing scripts to reproduce the exact input formats used in our experiments.  
    \item \textbf{Experimental setup:} Hyperparameters, optimization settings, and evaluation protocols are described in detail in Section~\ref{sec:eval}.  
    \item \textbf{Randomness:} All experiments are run with fixed random seeds in the scripts, to ensure consistent results.
    \item \textbf{Compute resources:} Our experiments are conducted on NVIDIA H100, RTX 4090 and RTX 5090 GPUs as described in Section~\ref{sec:eval}.
    \item \textbf{Limitations:} Some large-scale experiments (e.g., on 13B-parameter models) require access to high-end GPUs, which may limit reproducibility for groups without such resources.
\end{itemize}
We believe that with the released code, scripts, and dataset accessibility, other researchers will be able to reproduce our results and build upon our method.

\newpage
\bibliography{reference}
\bibliographystyle{iclr2026_conference}

\newpage
\appendix
\section{Appendix}
\subsection{The Use of LLMs}
Large language models (LLMs), such as ChatGPT, were used exclusively for language polishing and minor stylistic editing of the manuscript. All technical ideas, analyses, and experimental results were conceived, implemented, and verified by the authors. The authors carefully reviewed and validated all text to ensure accuracy.

\subsection{WSVD algorithm}
\label{sec:wsvd-appx}

\begin{algorithm}[h]
\caption{Weighted SVD Fine-tuning (WSVD) pseudo code}
\label{alg:wsvd}
\begin{algorithmic}[1]
\REQUIRE Calibration dataset $X$, model parameters $\{W\}$, rank $r$
\ENSURE Fine-tuned low-rank factors $\{A, B\}$

\FOR{each sample $x_i \in X$}
    \STATE Compute forward pass and loss $\mathcal{L}(x_i)$
    \STATE Backpropagate to obtain gradients $\{\nabla_W \mathcal{L}\}$
    \STATE Accumulate importance weights $F \gets \sum_X (\nabla_W \mathcal{L})^2$
\ENDFOR

\FOR{each weight matrix $W \in \{W_Q, W_K, W_V, \dots\}$}
    \STATE Compute SVD: $W_h \approx A_h B_h$, with $A_h \in \mathbb{R}^{m \times r}, B_h \in \mathbb{R}^{r \times n}$
    \STATE Define weighted loss:
    \[
        \mathcal{L}_{\text{WSVD}}(A_h,B_h)
        = \bigl\| F_h^{1/2} \odot (W_h - A_h B_h) \bigr\|_F^2
    \]
    where $F_h, W_h, A_h$ and $B_h$ is each head's importance weight, weight, decomposed matrices.
    \STATE Locally fine-tune $A_h, B_h$ using $\mathcal{L}_{\text{WSVD}}$
\ENDFOR

\RETURN Fine-tuned low-rank factors $\{A_h, B_h\}$ for all matrices
\end{algorithmic}
\end{algorithm}

\subsection{Detailed Results of WSVD-noQ}
\label{sec:noq_appx}
We inlcude the detailed results of WSVD-noQ and other baselines in Table~\ref{tab:full-svd-result}.
\begin{table*}[h]
    \newcommand{\bc}{\cellcolor{blue!10}}
    \newcommand{\accval}[1]{\fontsize{9.8pt}{10pt}\selectfont #1}
    \newcommand{\accvalours}[1]{\fontsize{9.75pt}{10pt}\selectfont \textbf{#1}}
    \centering
    \small
    \caption{Accuracy evaluation of different methods under FP16.}
    \resizebox{\linewidth}{!}{ %
    \begin{tabular}{l|l|c|c|c|c|c|c|c|c|c|c|c}
    \toprule
    
    \multirow{2}{*}{Acc.} & \multirow{2}{*}{Method} & \multicolumn{5}{c|}{ScienceQA-IMG $\uparrow$} & \multicolumn{5}{c|}{SEED-Bench $\uparrow$} & \multirow{2}{*}{Avg. $\uparrow$ } \\
    \cmidrule{3-12}
    & & $\rho_1: 90\%$ & $\rho_1: 80\%$ & $\rho_1: 70\%$ & $\rho_1: 60\%$ & $\rho_1: 50\%$ 
    & $\rho_1: 90\%$ & $\rho_1: 80\%$ & $\rho_1: 70\%$ & $\rho_1: 60\%$ & $\rho_1: 50\%$ \\
    
    \midrule
    
    \multirow{5.75}{*}{\rotatebox{90}{\makecell{LLaVA-v1.5\\7B}}}   

    & ASVD  
        & \accval{49.93\%} & \accval{50.12\%} & \accval{47.10\%} & \accval{36.69\%} & \accval{19.19\%} 
        & \accval{54.27\%} & \accval{53.53\%} & \accval{48.35\%} & \accval{37.17\%} & \accval{24.17\%} 
        & \accval{42.05\%}\\
    & SVD-LLM   
        & \accval{65.44\%} & \accval{63.71\%} & \accval{61.92\%} & \accval{57.41\%} & \accval{55.53\%}
        & \accval{57.89\%} & \accval{57.50\%} & \accval{55.33\%} & \accval{54.64\%} & \accval{55.31\%} 
        & \accval{58.47\%}\\
    & QSVD-noQ   
        & \accval{67.72\%} & \accval{\textbf{68.22\%}} & \accval{67.08\%} & \accval{65.05\%} & \accval{62.37\%} 
        & \accval{59.84\%} & \accval{59.07\%} & \accval{59.78\%} & \accval{59.00\%} & \accval{58.23\%} 
        & \accval{62.64\%}\\
    & \bc\textbf{WSVD-noQ} 
        & \bc\accval{\textbf{68.17\%}} & \bc\accval{67.72\%} & \bc\accval{\textbf{67.28\%}} & \bc\accval{\textbf{65.89\%}} & \bc\accval{\textbf{65.49\%}} 
        & \bc\accval{\textbf{60.10\%}} & \bc\accval{\textbf{60.17\%}} & \bc\accval{\textbf{59.89\%}} & \bc\accval{\textbf{60.18\%}} & \bc\accval{\textbf{60.46\%}} 
        & \bc\accval{\textbf{63.54\%}}\\
    \cline{3-12}
    \rule{0pt}{2.75ex}
    & \textcolor{gray}{FP16} & \multicolumn{5}{c|}{\textcolor{gray}{Accuracy: 68.01\%}} & \multicolumn{5}{c|}{\textcolor{gray}{Accuracy: 60.18\%}} & \textcolor{gray}{\accval{64.10\%}}  \\
    
    \midrule
    
    \multirow{5.75}{*}{\rotatebox{90}{\makecell{LLaVA-v1.5\\13B}}} %LLaVA-v1.5-13B

    & ASVD   
        & \accval{71.39\%} & \accval{71.59\%} & \accval{70.00\%} & \accval{70.25\%} & \accval{69.51\%} 
        & \accval{61.92\%} & \accval{61.91\%} & \accval{61.54\%} & \accval{61.51\%} & \accval{60.71\%} 
        & \accval{66.03\%}\\
    & SVD-LLM   
        & \accval{71.05\%} & \accval{70.85\%} & \accval{70.30\%} & \accval{70.35\%} & \accval{70.30\%}
        & \accval{62.28\%} & \accval{62.34\%} & \accval{62.25\%} & \accval{62.08\%} & \accval{\textbf{63.01\%}} 
        & \accval{66.48\%}\\
    & QSVD-noQ   
        & \accval{71.89\%} & \accval{\textbf{71.99\%}} & \accval{71.49\%} & \accval{71.54\%} & \accval{71.39\%} 
        & \accval{\textbf{62.61\%}} & \accval{62.64\%} & \accval{\textbf{62.82\%}} & \accval{\textbf{62.63\%}} & \accval{62.52\%} 
        & \accval{67.15\%}\\
    & \bc\textbf{WSVD-noQ} 
        & \bc\accval{\textbf{71.99\%}} & \bc\accval{71.84\%} & \bc\accval{\textbf{72.53\%}} & \bc\accval{\textbf{71.59\%}} & \bc\accval{\textbf{71.44\%}} 
        & \bc\accval{62.52\%} & \bc\accval{\textbf{62.68\%}} & \bc\accval{62.38\%} & \bc\accval{62.37\%} & \bc\accval{62.37\%} 
        & \bc\accval{\textbf{67.17\%}}\\
    \cline{3-12}
    \rule{0pt}{2.75ex}
    & \textcolor{gray}{FP16} & \multicolumn{5}{c|}{\textcolor{gray}{Accuracy: 71.83\%}} & \multicolumn{5}{c|}{\textcolor{gray}{Accuracy: 62.53\%}} & \textcolor{gray}{\accval{67.18\%}}\\

    \midrule
    
    \multirow{5.75}{*}{\rotatebox{90}{\makecell{LLaVA-Next\\7B}}} %LLaVA-v1.5-13B

    & ASVD   
        & \accval{64.20\%} & \accval{63.36\%} & \accval{62.07\%} & \accval{60.19\%} & \accval{55.28\%} 
        & \accval{67.38\%} & \accval{66.96\%} & \accval{66.24\%} & \accval{65.13\%} & \accval{61.52\%} 
        & \accval{63.23\%}\\
    & SVD-LLM   
        & \accval{68.27\%} & \accval{67.92\%} & \accval{66.58\%} & \accval{66.39\%} & \accval{65.54\%}
        & \accval{68.50\%} & \accval{68.31\%} & \accval{67.65\%} & \accval{67.45\%} & \accval{66.28\%} 
        & \accval{67.29\%}\\
    & QSVD-noQ   
        & \accval{\textbf{70.10\%}} & \accval{69.16\%} & \accval{69.01\%} & \accval{\textbf{68.27\%}} & \accval{66.19\%} 
        & \accval{68.86\%} & \accval{68.95\%} & \accval{68.44\%} & \accval{67.98\%} & \accval{67.27\%} 
        & \accval{68.42\%}\\
    & \bc\textbf{WSVD-noQ} 
        & \bc\accval{69.81\%} & \bc\accval{\textbf{69.56\%}} & \bc\accval{\textbf{69.36\%}} & \bc\accval{68.22\%} & \bc\accval{\textbf{67.87\%}}
        & \bc\accval{\textbf{69.18\%}} & \bc\accval{\textbf{69.27\%}} & \bc\accval{\textbf{69.15\%}} & \bc\accval{\textbf{69.16\%}} & \bc\accval{\textbf{68.59\%}} 
        & \bc\accval{\textbf{69.02\%}}\\
    \cline{3-12}
    \rule{0pt}{2.75ex}
    & \textcolor{gray}{FP16} & \multicolumn{5}{c|}{\textcolor{gray}{Accuracy: 69.60\%}} & \multicolumn{5}{c|}{\textcolor{gray}{Accuracy: 69.02\%}} & \textcolor{gray}{\accval{69.31\%}}\\

    \midrule
    
    \multirow{5.75}{*}{\rotatebox{90}{\makecell{LLaVA-Next\\13B}}} %LLaVA-v1.5-13B

    & ASVD   
        & \accval{71.24\%} & \accval{70.60\%} & \accval{71.44\%} & \accval{71.38\%} & \accval{69.81\%} 
        & \accval{70.88\%} & \accval{70.26\%} & \accval{70.01\%} & \accval{69.69\%} & \accval{69.01\%} 
        & \accval{70.43\%}\\
    & SVD-LLM   
        & \accval{72.53\%} & \accval{72.24\%} & \accval{71.74\%} & \accval{71.15\%} & \accval{70.55\%} 
        & \accval{70.76\%} & \accval{70.63\%} & \accval{70.25\%} & \accval{69.96\%} & \accval{69.58\%} 
        & \accval{70.94\%}\\
    & QSVD-noQ   
        & \accval{71.94\%} & \accval{72.14\%} & \accval{71.74\%} & \accval{72.14\%} & \accval{71.79\%} 
        & \accval{71.23\%} & \accval{71.02\%} & \accval{71.06\%} & \accval{70.92\%} & \accval{70.40\%} 
        & \accval{71.44\%}\\
    & \bc\textbf{WSVD-noQ} 
        & \bc\accval{\textbf{72.88\%}} & \bc\accval{\textbf{72.98\%}} & \bc\accval{\textbf{73.57\%}} & \bc\accval{\textbf{73.48\%}} & \bc\accval{\textbf{73.28\%}} 
        & \bc\accval{\textbf{71.29\%}} & \bc\accval{\textbf{71.17\%}} & \bc\accval{\textbf{71.25\%}} & \bc\accval{\textbf{70.95\%}} & \bc\accval{\textbf{70.81\%}} 
        & \bc\accval{\textbf{72.17\%}}\\
    \cline{3-12}
    \rule{0pt}{2.75ex}
    & \textcolor{gray}{FP16} & \multicolumn{5}{c|}{\textcolor{gray}{Accuracy: 73.23\%}} & \multicolumn{5}{c|}{\textcolor{gray}{Accuracy: 71.30\%}} & \textcolor{gray}{\accval{72.27\%}}\\

    \midrule
    
    \multirow{7}{*}{\rotatebox{90}{\makecell{SmolVLM\\2B}}}
    %\cline{2-14}
    %\rule{0pt}{2.25ex}
    %& & $\rho_1: 90\%$ & $\rho_1: 80\%$ & $\rho_1: 70\%$ & & & $\rho_1: 90\%$ & $\rho_1: 80\%$ & $\rho_1: 70\%$ & & \\
    
    & &
    \multicolumn{5}{l|}{%
      \begin{tabularx}{0.575\textwidth}{Y|Y|Y}
        $\rho_1: 90\%$ & $\rho_1: 80\%$ & $\rho_1: 70\%$
      \end{tabularx}
    }
    &
    \multicolumn{5}{l|}{%
      \begin{tabularx}{0.575\textwidth}{Y|Y|Y}
        $\rho_1: 90\%$ & $\rho_1: 80\%$ & $\rho_1: 70\%$
      \end{tabularx}
    } & \\    
    \cline{2-13}
    \rule{0pt}{2.75ex}
    & ASVD  
        & 
        \multicolumn{5}{l|}{%
          \begin{tabularx}{0.575\textwidth}{Y|Y|Y}
            \accval{29.30\%} & \accval{3.97\%} & \accval{0.20\%}
          \end{tabularx}
        }
        & \multicolumn{5}{l|}{%
          \begin{tabularx}{0.575\textwidth}{Y|Y|Y}
            \accval{17.85\%} & \accval{1.50\%} & \accval{0.95\%}
          \end{tabularx}
        }
        & \accval{8.96\%} \\
    & SVD-LLM   
        & 
        \multicolumn{5}{l|}{%
          \begin{tabularx}{0.575\textwidth}{Y|Y|Y}
            \accval{40.06\%} & \accval{17.20\%} & \accval{3.82\%}
          \end{tabularx}
        }
        & 
        \multicolumn{5}{l|}{%
          \begin{tabularx}{0.575\textwidth}{Y|Y|Y}
            \accval{32.49\%} & \accval{15.89\%} & \accval{4.60\%}
          \end{tabularx}
        }
        & \accval{19.01\%}\\
    & QSVD-noQ   
        & 
        \multicolumn{5}{l|}{%
          \begin{tabularx}{0.575\textwidth}{Y|Y|Y}
            \accval{\textbf{77.00\%}} & \accval{62.77\%} & \accval{42.59\%}
          \end{tabularx}
        }
        & 
        \multicolumn{5}{l|}{%
          \begin{tabularx}{0.575\textwidth}{Y|Y|Y}
            \accval{64.80\%} & \accval{50.46\%} & \accval{36.24\%}
          \end{tabularx}
        }
        & \accval{55.64\%}\\
    & \bc\textbf{WSVD-noQ} 
        & 
        \multicolumn{5}{l|}{%
          \bc\begin{tabularx}{0.575\textwidth}{Y|Y|Y}
            \accval{76.30\%} & \accval{\textbf{71.74\%}} & \accval{\textbf{60.93\%}}
          \end{tabularx}
        }
        & 
        \multicolumn{5}{l|}{%
          \bc\begin{tabularx}{0.575\textwidth}{Y|Y|Y}
            \accval{\textbf{65.78\%}} & \accval{\textbf{63.29\%}} & \accval{\textbf{54.45\%}}
          \end{tabularx}
        }
        & \bc\accval{\textbf{65.42\%}}\\
    \cline{3-12}
    \rule{0pt}{2.75ex}
    & \textcolor{gray}{FP16} & \multicolumn{5}{c|}{\textcolor{gray}{Accuracy: 84.53\%}} & \multicolumn{5}{c|}{\textcolor{gray}{Accuracy: 68.47\%}} & \textcolor{gray}{\accval{76.53\%}} \\
    
    \bottomrule
    \end{tabular}
    }
    \label{tab:full-svd-result}
\end{table*}

\subsection{Comparison with DobiSVD}
\label{sec:dobi-appx}

For completeness, we additionally evaluate DobiSVD on LLaVA-v1.5 7B using the ScienceQA-IMG benchmark. 
Following the official implementation, DobiSVD is applied to the $Q,K,V$ matrices while leaving other linear layers unchanged. 
Calibration is performed using the same set of samples and initialized with the same random seed as in the main experiments to ensure fairness. 
As shown in Table~\ref{tab:dobisvd}, WSVD-noQ achieves higher accuracy than DobiSVD under the same compression ratio.

\begin{table*}[h]
    \newcommand{\bc}{\cellcolor{blue!10}}
    \newcommand{\accval}[1]{\fontsize{9.8pt}{10pt}\selectfont #1}
    \newcommand{\accvalours}[1]{\fontsize{9.75pt}{10pt}\selectfont \textbf{#1}}
    \centering
    \small
    \caption{Accuracy evaluation of different methods under FP16.}
    \resizebox{0.7\linewidth}{!}{ %
    \begin{tabular}{l|l|c|c|c|c|c}
    \toprule
    
    \multirow{2}{*}{Acc.} & \multirow{2}{*}{Method} & \multicolumn{5}{c}{ScienceQA-IMG $\uparrow$} \\
    \cmidrule{3-7}
    & & $\rho_1: 90\%$ & $\rho_1: 80\%$ & $\rho_1: 70\%$ & $\rho_1: 60\%$ & $\rho_1: 50\%$ 
     \\
    
    \midrule
    
    \multirow{5.75}{*}{\rotatebox{90}{\makecell{LLaVA-v1.5\\7B}}}   
    & DobiSVD 
        & \accval{67.19\%} & \accval{60.94\%} & \accval{59.38\%} & \accval{56.64\%} & \accval{54.69\%} \\
        
    & ASVD  
        & \accval{49.93\%} & \accval{50.12\%} & \accval{47.10\%} & \accval{36.69\%} & \accval{19.19\%} \\
    & SVD-LLM   
        & \accval{65.44\%} & \accval{63.71\%} & \accval{61.92\%} & \accval{57.41\%} & \accval{55.53\%} \\
    & QSVD-noQ   
        & \accval{67.72\%} & \accval{\textbf{68.22\%}} & \accval{67.08\%} & \accval{65.05\%} & \accval{62.37\%} \\
    & \bc\textbf{WSVD-noQ} 
        & \bc\accval{\textbf{68.17\%}} & \bc\accval{67.72\%} & \bc\accval{\textbf{67.28\%}} & \bc\accval{\textbf{65.89\%}} & \bc\accval{\textbf{65.49\%}} \\
    \cline{3-7}
    \rule{0pt}{2.75ex}
    & \textcolor{gray}{FP16} & \multicolumn{5}{c}{\textcolor{gray}{Accuracy: 68.01\%}}  \\
    \bottomrule
    \end{tabular}
    }
    \label{tab:dobisvd}
\end{table*}

\subsection{Supplementary Results on More VLMs}
\label{sec:vlms-appx}
We additionally apply WSVD to Qwen-VL 7B~\citep{Qwen-VL} and Molmo-7B-O~\citep{deitke2025molmo} to further examine the generality of WSVD beyond the LLaVA family and SmolVLM. We follow exactly the same experimental setting as Section~\ref{sec:eval} and evaluate the FP16 and SVD compressed models on ScienceQA-IMG and SEED-Bench. As shown in Table~\ref{tab:more-vlms-result}, WSVD-noQ consistently outperforms all SVD-based baselines (ASVD and SVDLLM) across all singular-value ratios, and it also matches or slightly improves over the FP16 model. For example, on ScienceQA, WSVD-noQ improves over SVDLLM and ASVD by up to 3–5\%, and on SEED-Bench it yields the best/comparable accuracy among all compressed variants at every ratio. These results indicate that WSVD transfers well to VLMs with different vision–language fusion designs, supporting the general applicability of our method.

\begin{table*}[h]
    \newcommand{\bc}{\cellcolor{blue!10}}
    \newcommand{\accval}[1]{\fontsize{9.8pt}{10pt}\selectfont #1}
    \newcommand{\accvalours}[1]{\fontsize{9.75pt}{10pt}\selectfont \textbf{#1}}
    \centering
    \small
    \caption{Accuracy evaluation of different methods under FP16 on Qwen-VL 7B and Molmo-7B-O.}
    \resizebox{\linewidth}{!}{ %
    \begin{tabular}{l|l|c|c|c|c|c|c|c|c|c}
    \toprule
    
    \multirow{2}{*}{Acc.} & \multirow{2}{*}{Method} 
        & \multicolumn{4}{c|}{ScienceQA-IMG $\uparrow$} 
        & \multicolumn{4}{c|}{SEED-Bench $\uparrow$} 
        & \multirow{2}{*}{Avg. $\uparrow$ } \\
    \cmidrule{3-10}
    & & $\rho_1: 90\%$ & $\rho_1: 80\%$ & $\rho_1: 70\%$ & $\rho_1: 60\%$ 
      & $\rho_1: 90\%$ & $\rho_1: 80\%$ & $\rho_1: 70\%$ & $\rho_1: 60\%$ \\
    
    \midrule
    
    % ================== Qwen-VL-7B ==================
    \multirow{5.75}{*}{\rotatebox{90}{\makecell{Qwen-VL\\7B}}}   

    & ASVD  
        & \accval{63.86\%} & \accval{63.11\%} & \accval{60.54\%} & \accval{59.49\%} 
        & \accval{61.67\%} & \accval{61.00\%} & \accval{59.94\%} & \accval{58.69\%} 
        & \accval{61.04\%}\\
    & SVD-LLM   
        & \accval{65.29\%} & \accval{65.29\%} & \accval{64.55\%} & \accval{65.29\%}
        & \accval{62.99\%} & \accval{62.80\%} & \accval{62.69\%} & \accval{62.51\%} 
        & \accval{63.93\%}\\
    & QSVD-noQ   
        & \accval{66.78\%} & \accval{66.24\%} & \accval{66.68\%} & \accval{65.20\%} 
        & \accval{63.25\%} & \accval{63.00\%} & \accval{62.06\%} & \accval{61.26\%} 
        & \accval{64.31\%}\\
    & \bc\textbf{WSVD-noQ} 
        & \bc\accval{\textbf{68.77\%}} & \bc\accval{\textbf{68.12\%}} & \bc\accval{\textbf{67.23\%}} & \bc\accval{\textbf{65.49\%}} 
        & \bc\accval{\textbf{63.32\%}} & \bc\accval{\textbf{63.53\%}} & \bc\accval{\textbf{63.60\%}} & \bc\accval{\textbf{63.68\%}} 
        & \bc\accval{\textbf{65.47\%}}\\
    \cline{3-10}
    \rule{0pt}{2.75ex}
    & \textcolor{gray}{FP16} 
        & \multicolumn{4}{c|}{\textcolor{gray}{Accuracy: 68.32\%}} 
        & \multicolumn{4}{c|}{\textcolor{gray}{Accuracy: 63.52\%}} 
        & \textcolor{gray}{\accval{65.92\%}}  \\
    
    \midrule
    
    % ================== Molmo-7B-O ==================
    \multirow{5.75}{*}{\rotatebox{90}{\makecell{Molmo\\7B-O}}}

    & ASVD   
        & \accval{94.99\%} & \accval{95.19\%} & \accval{94.10\%} & \accval{93.06\%} 
        & \accval{74.42\%} & \accval{74.21\%} & \accval{73.73\%} & \accval{73.62\%} 
        & \accval{84.17\%}\\
    & SVD-LLM   
        & \accval{95.09\%} & \accval{95.09\%} & \accval{94.99\%} & \accval{\textbf{95.09\%}}
        & \accval{\textbf{74.68\%}} & \accval{74.46\%} & \accval{74.23\%} & \accval{74.11\%} 
        & \accval{84.72\%}\\
    & QSVD-noQ   
        & \accval{95.54\%} & \accval{94.99\%} & \accval{94.59\%} & \accval{93.85\%} 
        & \accval{74.47\%} & \accval{74.44\%} & \accval{74.41\%} & \accval{74.37\%} 
        & \accval{84.58\%}\\
    & \bc\textbf{WSVD-noQ} 
        & \bc\accval{\textbf{95.59\%}} & \bc\accval{\textbf{95.49\%}} & \bc\accval{\textbf{95.34\%}} & \bc\accval{\textbf{95.09\%}} 
        & \bc\accval{74.61\%} & \bc\accval{\textbf{74.52\%}} & \bc\accval{\textbf{74.48\%}} & \bc\accval{\textbf{74.38\%}} 
        & \bc\accval{\textbf{84.94\%}}\\
    \cline{3-10}
    \rule{0pt}{2.75ex}
    & \textcolor{gray}{FP16} 
        & \multicolumn{4}{c|}{\textcolor{gray}{Accuracy: 95.78\%}} 
        & \multicolumn{4}{c|}{\textcolor{gray}{Accuracy: 74.74\%}} 
        & \textcolor{gray}{\accval{85.26\%}}\\
    \bottomrule
    \end{tabular}
    }
    \label{tab:more-vlms-result}
    \vspace{-10pt}
\end{table*}

\subsection{Supplementary Results on More Datasets}
\label{sec:datasets-appx}
To further assess generalization, we additionally evaluate LLaVA-Next 13B with WSVD-noQ on two additional benchmarks: HRBench-4K (4K high-resolution images)~\citep{wang2025divide}, and OCRBench (text-centric images)~\citep{liu2024ocrbench}. Following Section~\ref{sec:eval}, we reuse the same 256-sample calibration set drawn from the ScienceQA training set and keep all other settings identical, while sweeping the parameter ratios $\rho_1$. As summarized in Table~\ref{tab:more-datasets-result}, WSVD-noQ consistently matches or outperforms all baselines across nearly all ratios on these datasets, despite being calibrated only once on the ScienceQA training set. These results indicate that WSVD generalizes well across tasks and datasets. Moreover, WSVD’s decoding speedup is independent of the evaluation dataset: once the model is calibrated and compressed, runtime is determined solely by the resulting model size and context length, so a fixed compressed model yields essentially the same speedup across benchmarks.

\begin{table*}[h]
    \newcommand{\bc}{\cellcolor{blue!10}}
    \newcommand{\accval}[1]{\fontsize{9.8pt}{10pt}\selectfont #1}
    \newcommand{\accvalours}[1]{\fontsize{9.75pt}{10pt}\selectfont \textbf{#1}}
    \centering
    \small
    \caption{Accuracy evaluation of different methods under FP16 on OCRBench and HRBench-4K.}
    \resizebox{\linewidth}{!}{ %
    \begin{tabular}{l|l|c|c|c|c|c|c|c|c|c|c|c}
    \toprule
    
    \multirow{2}{*}{Acc.} & \multirow{2}{*}{Method} & \multicolumn{5}{c|}{OCRBench $\uparrow$} & \multicolumn{5}{c|}{HRBench-4K $\uparrow$} & \multirow{2}{*}{Avg. $\uparrow$ } \\
    \cmidrule{3-12}
    & & $\rho_1: 90\%$ & $\rho_1: 80\%$ & $\rho_1: 70\%$ & $\rho_1: 60\%$ & $\rho_1: 50\%$ 
    & $\rho_1: 90\%$ & $\rho_1: 80\%$ & $\rho_1: 70\%$ & $\rho_1: 60\%$ & $\rho_1: 50\%$ \\
    
    \midrule
    
     \multirow{5.75}{*}{\rotatebox{90}{\makecell{LLaVA-Next\\13B}}}   

    & ASVD  
        & \accval{53.10\%} & \accval{52.50\%} & \accval{51.30\%} & \accval{50.50\%} & \accval{47.70\%} 
        & \accval{44.00\%} & \accval{44.13\%} & \accval{43.88\%} & \accval{43.00\%} & \accval{42.00\%} 
        & \accval{47.21\%}\\
    & SVD-LLM   
        & \accval{52.10\%} & \accval{51.90\%} & \accval{51.00\%} & \accval{49.90\%} & \accval{48.20\%}
        & \accval{43.00\%} & \accval{44.25\%} & \accval{42.62\%} & \accval{43.75\%} & \accval{43.25\%} 
        & \accval{47.00\%}\\
    & QSVD-noQ   
        & \accval{52.80\%} & \accval{52.40\%} & \accval{52.40\%} & \accval{51.40\%} & \accval{\textbf{48.90\%}} 
        & \accval{44.25\%} & \accval{\textbf{44.88\%}} & \accval{43.88\%} & \accval{43.37\%} & \accval{42.88\%} 
        & \accval{47.72\%}\\
    & \bc\textbf{WSVD-noQ} 
        & \bc\accval{\textbf{53.30\%}} & \bc\accval{\textbf{53.30\%}} & \bc\accval{\textbf{53.50\%}} & \bc\accval{\textbf{52.20\%}} & \bc\accval{48.70\%} 
        & \bc\accval{\textbf{46.13\%}} & \bc\accval{\textbf{44.88\%}} & \bc\accval{\textbf{44.50\%}} & \bc\accval{\textbf{44.88\%}} & \bc\accval{\textbf{44.50\%}} 
        & \bc\accval{\textbf{48.59\%}}\\
    \cline{3-12}
    \rule{0pt}{2.75ex}
    & \textcolor{gray}{FP16} 
        & \multicolumn{5}{c|}{\textcolor{gray}{Accuracy: 53.30\%}} 
        & \multicolumn{5}{c|}{\textcolor{gray}{Accuracy: 45.63\%}} 
        & \textcolor{gray}{\accval{49.47\%}}  \\
    \bottomrule
    \end{tabular}
    }
    \label{tab:more-datasets-result}
    \vspace{-10pt}
\end{table*}

% \subsection{\textcolor{blue}{Latency Comparison of Different Rank Ratios}}
% \label{sec:latency-rank-ratio-appx}
% \textcolor{blue}{
% Using the same setting as Section~\ref{sec:eval-system}, we vary the rank ratio $\rho_2 \in \{90\%, 70\%, 50\%\}$ for WSVD-noQ and measure the latency on RTX~4090 and RTX~5090 GPUs. Table~\ref{tab:rank-latency} summarizes the impact of rank on decoding latency. Smaller $\rho_2$ (lower ranks) consistently leads to lower latency on both devices, showing that WSVD’s fused kernel effectively converts reduced ranks into practical decoding speedups.
% }
% \begin{table}[h]
%     \centering
%     \small
%     \caption{Decoding latency (ms) under different $\rho_2$.}
%     \begin{tabular}{lcccc}
%         \toprule
%         GPU & Flash Decoding & $\rho_2:90\%$ & $\rho_2:70\%$ & $\rho_2:50\%$ \\
%         \midrule
%         4090 & 2.92 & 2.83 & 2.53 & 1.66 \\
%         5090 & 2.14 & 1.87 & 1.75 & 1.15 \\
%         \bottomrule
%     \end{tabular}
%     \label{tab:rank-latency}
% \end{table}

\subsection{Latency Comparison of Full and Per-Head SVD}
\label{sec:latency-appx}
We further compare the decoding latency of applying SVD to the full QKV matrices versus adopting WSVD's fine-grained per-head SVD. 
To enable this comparison, we minimally modify our kernel to support reconstruction with larger matrix sizes under the full SVD setting (as discussed in Section~\ref{sec:wsvd}), while still fusing the reconstruction with flash decoding. 
This variant is denoted as “W/o per-head.” 
Both approaches are evaluated under the same $\rho_2$, ensuring equal overall cache size, with batch size, sequence length and other settings kept identical to the setup described above.

\begin{table}[h]
  \centering
  \begin{minipage}[t]{0.48\textwidth}
    \centering
    \caption{Decoding latency on RTX 4090.}
    \resizebox{\linewidth}{!}{
      \begin{tabular}{c|c|c|c}
        \toprule
        $\rho_2$ & W/o per-head & WSVD-noQ & Speedup \\
        \midrule
        90\% & 51.31 & 2.83 & $18.1\times$ \\
        80\% & 46.03 & 2.60 & $17.7\times$ \\
        70\% & 39.46 & 2.53 & $15.6\times$ \\
        60\% & 33.54 & 2.25 & $14.9\times$ \\
        50\% & 28.37 & 1.66 & $17.1\times$ \\
        \bottomrule
      \end{tabular}
    }
    \label{tab:latency_4090}
  \end{minipage}
  \hfill
  \begin{minipage}[t]{0.48\textwidth}
    \centering
    \caption{Decoding latency on RTX 5090.}
    \resizebox{\linewidth}{!}{
      \begin{tabular}{c|c|c|c}
        \toprule
        $\rho_2$ & W/o per-head & WSVD-noQ & Speedup \\
        \midrule
        90\% & 40.40 & 1.87 & $21.6\times$ \\
        80\% & 35.99 & 1.77 & $20.3\times$ \\
        70\% & 31.14 & 1.75 & $17.8\times$ \\
        60\% & 26.85 & 1.56 & $17.2\times$ \\
        50\% & 21.44 & 1.15 & $18.6\times$ \\
        \bottomrule
      \end{tabular}
    }
    \label{tab:latency_5090}
  \end{minipage}
\end{table}

As shown in Tables~\ref{tab:latency_4090} and \ref{tab:latency_5090}, WSVD-noQ consistently achieves more than an order-of-magnitude speedup over the full-matrix SVD variant (“W/o per-head”) across all compression ratios $\rho_2$. 
On RTX 4090, the speedup ranges from $14.9\times$ to $18.1\times$, while on RTX 5090 it further increases to $17.2\times$–$21.6\times$. 
These results confirm that per-head SVD substantially reduces reconstruction overhead and I/O traffic, enabling efficient decoding.

\subsection{Training Cost of WSVD}
\label{sec:cost-appx}
WSVD first applies SVDLLM's whitening method~\citep{wang2024svd} to per-head weight matrices before performing SVD, then uses QSVD's importance-score-based rank allocation~\citep{wang2025qsvd} to truncate the model, and subsequently performs lightweight, Fisher-information-based local fine-tuning and local quantization-aware training on the truncated low-rank weights to better preserve the most sensitive weight elements and to mitigate the degradation of per-head SVD and low-precision inference.

\begin{table}[h]
    \centering
    \small
    \caption{Calibration time breakdown for QSVD and WSVD.}
    \begin{tabular}{l c | l c}
        \toprule
        \multicolumn{2}{c|}{QSVD} & \multicolumn{2}{c}{WSVD} \\
        \cmidrule(r){1-2} \cmidrule(l){3-4}
        Step & Time & Step & Time \\
        \midrule
        Input calibration                  & 1 min         & Input calibration            & 6 min        \\
        SVD on all layers              & 2 min 15 s    & SVD on all layers         & 1 min        \\
        Gradient collection \& rank allocation & 10 min 12 s & Gradient collection \& rank allocation & 13 min 40 s \\
        SVD results fusion                & 30 s          & Local FT                    & 9 min        \\
        $\beta$ tuning \& quantization    & 82 min        & Local QAT \& quantization   & 8 min        \\
        \midrule
        Total calibration time            & 96 min        & Total calibration time      & 38 min       \\
        \bottomrule
    \end{tabular}
    \label{tab:calibration_time_qsvd_wsvd}
\end{table}

We quantify the computational overhead of WSVD's local fine-tuning and QAT and compare it with QSVD on LLaVA-1.5 13B. We extract QSVD's reported training time from its OpenReview page and, for fairness, benchmark WSVD on the same GPU type (A100). As shown in Table~\ref{tab:calibration_time_qsvd_wsvd}, QSVD requires \textbf{96 minutes $\approx$ 1.6 A100 GPU-hours}, whereas WSVD takes only \textbf{38 minutes $\approx$ 0.63 A100 GPU-hours}, about \textbf{2.5$\times$ less tuning time}. The peak GPU memory usage of WSVD's local FT and QAT stages on LLaVA-1.5 13B is only \textbf{15 GB}, since we only perform local updates on low-rank weights rather than end-to-end fine-tuning and do not store full intermediate activations, so the whole procedure fits comfortably on a single A100-80GB. For additional context, the official LLaVA-1.5 report~\citep{liu2024improved} states that training the 13B model requires at least \textbf{204 A100 GPU-hours}. Thus, WSVD's tuning cost is only a small fraction of the original training cost, while still delivering practical decoding speedups, indicating that the efficiency gains comfortably justify the modest local fine-tuning and QAT overhead.

\end{document}